\documentclass[letterpaper]{article} 
\usepackage{aaai2027}  
\usepackage[hyphens]{url}  
\usepackage{graphicx} 
\usepackage{natbib}  
\usepackage{caption} 
\usepackage{algorithm}
\usepackage{algorithmic}
\usepackage{times}
\usepackage{latexsym}
\usepackage[T1]{fontenc}
\usepackage[utf8]{inputenc}
\usepackage{microtype}
\usepackage{inconsolata}
\usepackage{graphicx}
\usepackage{booktabs}
\usepackage{amsmath}
\usepackage{amssymb}
\usepackage{enumitem}
\usepackage{xcolor}
\usepackage{listings}
\usepackage[hyphens]{url}  
\usepackage{graphicx} 
\usepackage{natbib}  
\usepackage{caption} 

\usepackage{newfloat}
\usepackage{listings}
\DeclareCaptionStyle{ruled}{labelfont=normalfont,labelsep=colon,strut=off} 
\floatstyle{ruled}
\newfloat{listing}{tb}{lst}{}
\floatname{listing}{Listing}

\usepackage{booktabs}

\title{Internalizing Academic Writing Workflows for Introduction Generation via Struct-Aware Policy Learning}

\author{
    Meicong Zhang\textsuperscript{\rm 1},
    Tiancheng Su\textsuperscript{\rm 1},
    Jiahao Cheng\textsuperscript{\rm 1},
    Guoxiu He\textsuperscript{\rm 1}\corresponding,\\
    Xinqi Tao\textsuperscript{\rm 2},
    Dejia Song\textsuperscript{\rm 2}
}

\affiliations{
    \textsuperscript{\rm 1}School of Economics and Management, East China Normal University, Shanghai, China\\
    \textsuperscript{\rm 2}Xiaohongshu Inc., Shanghai, China\\
    \{mczhang,tcsu,chengjiahao\}@stu.ecnu.edu.cn,
    gxhe@fem.ecnu.edu.cn,
    \{taoxinqi,dejiasong\}@xiaohongshu.com
}

\begin{document}

\maketitle

\begin{abstract}
Generating a rigorous paper introduction with large language models (LLMs) remains challenging, since it requires coordinating background, gap identification, method and contribution within a coherent narrative. Existing solutions externalize this process as multi-stage prompts or agent workflows which are expensive and vulnerable to cross-stage drift. We propose StructPO, a struct-aware policy learning framework that internalizes the entire multi-stage writing workflow into a single-pass policy controlled by explicit stage tokens. StructPO introduces struct-aware credit assignment to decouple local stage quality from global coherence and refinement-guided optimization to internalize revision behavior into the first-pass policy. Experiments show that StructPO improves semantic alignment, structural rationality and inference efficiency over workflow-based baselines, generalizes to out-of-domain settings, and remains competitive with GPT-5.1 in human evaluation when scaled to Qwen3-32B. These results show that internalizing academic writing workflows through fine-grained policy optimization offers a viable alternative to costly external orchestration. 
\end{abstract}


\section{Introduction}
\label{sec:intro}

LLMs have recently been applied to many parts of scientific workflows, including literature understanding, hypothesis generation, experimental support and manuscript drafting \citep{zhang2025exploring,DBLP:journals/corr/abs-2408-06292,garikaparthi-etal-2025-mir,li2025scilitllm,Langley_2024,wang2023scientific,yang-etal-2024-large-language,zhao-etal-2025-abgen,boiko2023autonomous}. 
In academic paper writing, the introduction is the section where a study first presents its motivation, novelty and contribution to readers. 
It must connect the study to prior work, identify limitations in existing approaches, motivate the proposed method and preview the paper's contributions in a coherent rhetorical sequence. 
We therefore focus on introduction generation, which combines background construction, gap identification, method positioning and contribution framing within a single section.

Existing LLM-based academic writing systems have made progress on literature reviews, survey generation, and general manuscript assistance \citep{NEURIPS2024_d07a9fc7,DBLP:journals/corr/abs-2503-04629,DBLP:journals/corr/abs-2504-18765}. 
To handle long-form writing, these systems often rely on multi-stage prompts or agent workflows that separate outlining, drafting, revising and polishing. 
However, such external orchestration is usually handcrafted, incurs repeated context replay and may introduce cross-stage drift, where later stages deviate from earlier plans or repeat incompatible rhetorical content.


A natural alternative is to internalize the writing workflow into the model, enabling structured introduction generation in a single pass. 
Yet supervised fine-tuning provides limited direct feedback on long-range rhetorical organization, while standard policy optimization methods such as PPO and GRPO \citep{schulman2017proximalpolicyoptimizationalgorithms,shao2024grpo} typically assign one scalar reward to the whole output. 
For structured writing, this coarse feedback makes it difficult to identify which rhetorical section succeeds or fails, leading to noisy credit assignment and unstable optimization.

In this work, we propose StructPO, a struct-aware policy optimization framework for academic introduction generation. StructPO internalizes multi-stage academic writing workflows into a single-pass policy controlled by explicit stage tokens. Specifically, an introduction is represented as eight structural units corresponding to four rhetorical sections: background, problem, method and contributions. Each section consists of an outline and corresponding content. This representation exposes stage boundaries during training and avoids external workflow orchestration during inference.

To optimize this structured generation process, StructPO introduces a Struct-aware Relative Advantage (SRA) estimator. Rather than assigning the same advantage to all tokens in an introduction, SRA computes stage-level advantages and combines them with a global alignment signal. This allows LLMs to reinforce high-quality behavior within each rhetorical stage while still preserving document-level coherence. In addition, StructPO incorporates refinement-guided optimization. During training, LLMs generate draft--revision trajectories and drafts that are substantially improved by their revisions receive an additional penalty.

We evaluate StructPO on 1,176 ACL 2025 papers and compare it against prompt-based systems, workflow-based baselines and a suite of strong closed-source LLMs including GPT-4o, GPT-5.1, GPT-5.5, Claude Opus 4.8, and Gemini-3.1-Pro-Preview. StructPO achieves near-perfect structure scores, strong semantic similarity and substantially lower inference overhead. When scaled to a Qwen3-32B backbone, StructPO further becomes competitive with GPT-5.1 in blind human evaluation and matches or surpasses several closed-source LLMs on automated metrics. Further analyses show that StructPO mitigates length collapse, reduces reward hacking and transfers to CVPR papers.

Our contributions are summarized as follows:
\begin{itemize}
    \item We formulate academic introduction generation as a single-pass structured policy and propose StructPO to internalize multi-stage writing workflows.
    \item We develop Struct-aware Relative Advantage estimation and a refinement-guided penalty to support stage-level credit assignment and training-time distillation of revision behavior.
     \item We provide empirical evaluations on ACL and CVPR papers, showing that StructPO improves structural control and inference efficiency over prompt-based and workflow-based baselines. When scaled to a Qwen3-32B backbone, StructPO surpasses all evaluated closed-source LLMs on automated metrics and remains competitive with GPT-5.1 and Claude Opus 4.8 under judge-based and human evaluation.
\end{itemize}

\section{Related Work}
\label{sec:related}

\subsection{LLM-based Scientific Writing Systems}
LLM-based scientific writing systems support idea generation, literature review, experimental design, result analysis, citation assistance and drafting \citep{su-etal-2025-many,wu2023autogen,li-etal-2025-chain-ideas,zimmermann2024leveraging,agarwal2024litllm,schmidgall2025agentlaboratoryusingllm,ge-etal-2021-baco,wang2021autocite}. AutoSurvey and SurveyForge further show the effectiveness of workflow decomposition for literature review \citep{NEURIPS2024_d07a9fc7,DBLP:journals/corr/abs-2503-04629}.

However, introduction writing requires tighter rhetorical coordination among background construction, gap articulation, method positioning and contribution framing. Recent systems such as ResearchAgent support broader research-assistance capabilities \citep{baek-etal-2025-researchagent}, but still largely rely on multi-stage orchestration \citep{DBLP:journals/corr/abs-2504-18765}. In contrast, we aim to internalize this writing logic within the policy.

\subsection{Step-Level Policy Optimization}

Reinforcement learning has become a key paradigm for aligning LLMs with human preferences and task-specific objectives \citep{ouyang2022training}. Policy-gradient methods such as REINFORCE \citep{williams1992simple} and PPO \citep{schulman2017proximalpolicyoptimizationalgorithms} optimize policies from scalar rewards, while GRPO \citep{shao2024grpo} removes the critic by normalizing rewards within sampled groups.

However, trajectory-level rewards are often too coarse for long-horizon generation. Recent work therefore explores finer-grained credit assignment, including hierarchical grouping, Shapley-value decomposition, information-gain-based supervision and calibrated step-wise advantages \citep{feng2025groupingroup,li2026ssvpo,huang2026sketchvl,wang2026information,fei2025selfguidedprocessrewardoptimization}. These methods demonstrate the value of dense optimization signals, but they mainly target homogeneous reasoning tasks such as mathematics or coding.

Structured academic writing poses a different challenge because planning and realization stages are heterogeneous yet interdependent. Outlines require coverage and organization, whereas content paragraphs require semantic fidelity, length control, and rhetorical consistency. StructPO addresses this gap by assigning struct-aware advantages to different writing stages while maintaining global introduction-level coherence.

\section{Methodology}
\label{sec:methodology}

\begin{figure*}[t]
  \centering
  \includegraphics[width=0.9\textwidth]{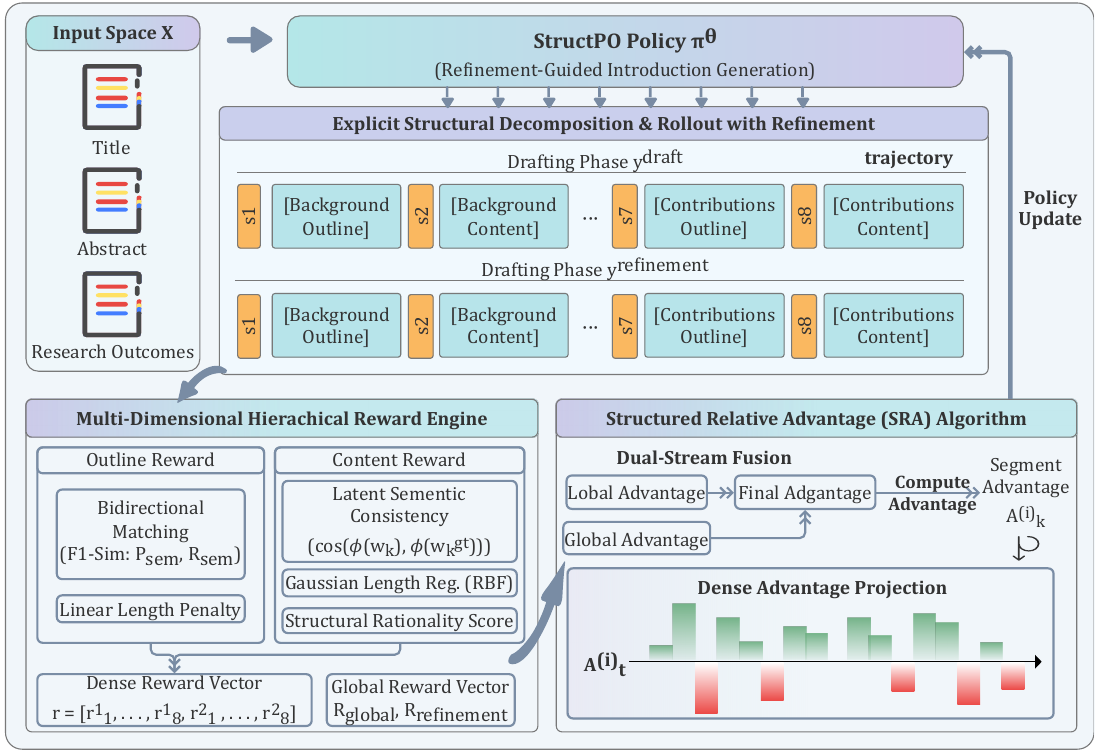}
    \caption{Overview of the StructPO training pipeline. Given paper metadata, the policy generates draft--revision trajectories with explicit stage tokens. A modality-aware reward engine scores outline and content stages, and SRA projects fused local-global advantages back to tokens for fine-grained policy optimization.}
  \label{fig:structpo}
\end{figure*}

We formulate introduction writing as a structured policy-learning problem. As shown in Figure~\ref{fig:structpo}, StructPO combines explicit stage decomposition, modality-aware stage rewards, and SRA-based dense optimization. Training uses draft--revision trajectories so the model acquires revision behavior during training while requiring only a single pass at inference time.

\subsection{Task Definition: Explicit Structural Decomposition}
\label{subsec:task_definition}

We formulate introduction generation as conditional sequence generation from input $x=\{x_{\text{tit}},x_{\text{abs}},x_{\text{exp}}\}$, where the three terms denote the title, abstract, and extracted experimental metadata. To reduce structural drift, we decompose each introduction into $K=8$ units defined by four rhetorical sections and two modalities:
\begin{equation}
\mathbb{S}=
\{\mathcal{P}_{bg},\mathcal{P}_{prob},\mathcal{P}_{meth},\mathcal{P}_{contrib}\}
\times
\{\mathcal{M}_{out},\mathcal{M}_{con}\}.
\end{equation}
Each stage is generated between explicit control tokens, and the full trajectory is
\begin{equation}
y=\bigoplus_{k=1}^{K}
\left(\langle s_k\rangle \oplus w_k \oplus \langle e_k\rangle\right),
\end{equation}
where $w_k=(w_{k,1},\ldots,w_{k,T_k})$ is the generated text of stage $k$. This grammar exposes stage boundaries for reward computation and enables single-pass structured generation at inference time.

\subsection{Rollout Strategy: Interactive Trajectory Generation}
\label{subsec:rollout}

To distill revision behavior into the policy, training uses draft--revision trajectories. For each input, the model first generates a draft $y^{\mathrm{d}}$ and then a revision $y^{\mathrm{r}}$, forming $\tau=[y^{\mathrm{d}},y^{\mathrm{r}}]=[u_1,\ldots,u_K,u_{K+1},\ldots,u_{2K}]$, where the first $K$ units belong to the draft phase and the remaining $K$ units to the revision phase. Following GRPO, we sample $N$ trajectories from $\pi_{\theta_{\text{old}}}$ for each input and compute relative advantages within the sampled group.

\subsection{Reward Design}
\label{subsec:reward}

StructPO scores each structural unit separately to provide dense supervision for both planning and content realization. For dense content stages, we apply Gaussian length regularization, $r_{\text{len}}^{\text{gau}}(w_k)=\exp[-(l(w_k)-L_k^{*})^2/(2\delta^2)]$, where $l(w_k)$ is the word count, $L_k^{*}$ is the target length, and $\delta$ is the bandwidth. We set $\delta=6$ for section-level rewards and $\delta=12$ for the global length reward. For outline stages, we use a linear sentence-count penalty, $r_{\text{len}}^{\text{lin}}(w_k)=1-|N_{\text{gen}}-N_{\text{ref}}|/\max(N_{\text{ref}},1)$.

For content stages, semantic fidelity is measured by embedding similarity, $r_{\text{sim}}(w_k)=\cos(\phi(w_k),\phi(w_k^{\text{gt}}))$, where $\phi(\cdot)$ is a pretrained dense encoder such as Qwen-Embedding~\citep{qwen3embedding}. For outline stages, we use bidirectional semantic matching. Given generated and reference outline sentence embeddings $\{s_i\}_{i=1}^{m}$ and $\{r_j\}_{j=1}^{n}$, with $M_{ij}=\cos(s_i,r_j)$, we compute
\begin{equation}
\begin{aligned}
P_{\text{sem}} &= \frac{1}{m}\sum_{i=1}^{m}\max_j M_{ij}, \qquad
R_{\text{sem}} = \frac{1}{n}\sum_{j=1}^{n}\max_i M_{ij},\\
r_{\text{match}} &= 
\frac{2P_{\text{sem}}R_{\text{sem}}}
{P_{\text{sem}}+R_{\text{sem}}+\varepsilon_0}.
\end{aligned}
\end{equation}

To enforce rhetorical purity, we train a DeBERTa-v3 classifier~\citep{he2021debertav3} to predict the intended rhetorical section of each sentence. For stage $k$, the structural rationality score is $r_{\text{struct}}(w_k)=1-C_{\text{mis}}/C_{\text{tot}}$, where $C_{\text{mis}}$ is the number of sentences assigned to the wrong section and $C_{\text{tot}}$ is the total number of sentences.

The final local reward depends on the modality of stage $k$:
\begin{equation}
r_k=
\begin{cases}
 r_{\text{match}}(w_k)\, r_{\text{len}}^{\text{lin}}(w_k),
 & \text{Outline},\\[3pt]
 \omega_1 r_{\text{sim}}(w_k) r_{\text{struct}}(w_k)
 + \omega_2 r_{\text{len}}^{\text{gau}}(w_k),
 & \text{Content}.
\end{cases}
\end{equation}
The reward vector $\mathbf{r}=[r_1,\ldots,r_{2K}]$ provides dense local supervision, while a global reward $R_{\text{global}}$ measures whole-introduction alignment with the reference.

\subsection{Struct-aware Relative Advantage (SRA) and Dense Optimization}
\label{subsec:sra_optimization}

Standard GRPO assigns a single trajectory-level signal to every token:
\begin{equation}
\hat{g}_{\text{GRPO}}
= \mathbb{E}\!\left[
\nabla_\theta \log \pi(y_t)
\cdot \frac{R(\tau)-\mu_{\text{glob}}}{\sigma_{\text{glob}}}
\right],
\end{equation}
where $R(\tau)=\sum_{k=1}^{2K} r_k$ is the cumulative trajectory reward, and $\mu_{\text{glob}}, \sigma_{\text{glob}}$ are the mean and standard deviation of $R(\tau)$ within the sampled group. For long-form generation, this formulation introduces attribution noise because rewards from unrelated sections affect every token, and the gradient variance grows with cross-section interference:
\begin{equation}
\mathrm{Var}(\hat{g}_{\text{GRPO}})
\propto
\mathrm{Var}(r_k) + \sum_{j\neq k}\mathrm{Var}(r_j).
\end{equation}
Consequently, a high-quality background paragraph can be penalized because of errors in the method paragraph.

To mitigate this issue, StructPO normalizes stage rewards and fuses local and global signals. We first map each local reward to a comparable scale:
\begin{equation}
r_{k,\text{norm}} = \frac{r_k}{r_k^{\max}},
\end{equation}
where $r_k^{\max}$ denotes the theoretical upper bound of $r_k$ derived from its component metrics. We then decouple credit assignment into two complementary streams. For trajectory $i$ and stage $k$, the \emph{local advantage} normalizes rewards within the same stage across the group,
\begin{equation}
A_{\text{local}}^{(i,k)} = \frac{r_k^{(i)} - \mu_k}{\sigma_k},
\end{equation}
where $\mu_k$ and $\sigma_k$ are the within-stage mean and standard deviation across the $N$ sampled trajectories. The \emph{global advantage} $A_{\text{global}}^{(i)}$ is defined analogously by normalizing $R_{\text{global}}^{(i)}$ within the group. The fused advantage is
\begin{equation}
\hat{A}_k^{(i)}
= \lambda\, A_{\text{global}}^{(i)}
+ (1-\lambda)\, A_{\text{local}}^{(i,k)},
\end{equation}
where $\lambda\in[0,1]$ controls the trade-off between global coherence and local stage quality.

To incorporate revision supervision, we define the stage-level refinement gap as
\begin{equation}
\Delta_k^{(i)}
= \max\!\bigl(0,\; r_k(\tau^{(i)}_{\mathrm{r}}) - r_k(\tau^{(i)}_{\mathrm{d}})\bigr),
\end{equation}
where $\tau^{(i)}_{\mathrm{d}}$ and $\tau^{(i)}_{\mathrm{r}}$ denote the draft and revision phases of trajectory $i$. The struct-aware relative advantage with refinement guidance is then
\begin{equation}
\mathcal{A}_k^{(i)}
= \hat{A}_k^{(i)}
- \mathbb{I}\!\bigl(\tau^{(i)}\in\text{Draft}\bigr)
\cdot \eta\, \frac{\Delta_k^{(i)}}{\sigma_k},
\end{equation}
where $\eta\geq 0$ is the refinement penalty coefficient and $\mathbb{I}(\cdot)$ is the indicator function. This term explicitly penalizes draft stages that are substantially improved by their revisions.

We then project stage-level advantages back to tokens:
\begin{equation}
\mathcal{A}_t^{(i)}
= \sum_{k=1}^{2K}
\mathbb{I}\!\bigl(t_{\text{start}}^{(i,k)}\le t \le t_{\text{end}}^{(i,k)}\bigr)
\cdot \mathcal{A}_k^{(i)},
\end{equation}
where $t_{\text{start}}^{(i,k)}$ and $t_{\text{end}}^{(i,k)}$ are the token positions immediately following $\langle s_k\rangle$ and immediately preceding $\langle e_k\rangle$ in trajectory $i$. The final objective is
\begin{equation}
\small
\begin{aligned}
\mathcal{L}_{\text{StructPO}}(\theta)
= {} & -\frac{1}{N}\sum_{i=1}^{N}\sum_{t=1}^{T_i}
\mathbb{I}\!\bigl(t\in\mathcal{S}_{\text{val}}^{(i)}\bigr) \\
& \cdot \min\!\Bigl(
\rho_t^{(i)}\mathcal{A}_t^{(i)},\;
\operatorname{clip}\!\bigl(\rho_t^{(i)},1-\epsilon_c,1+\epsilon_c\bigr)\mathcal{A}_t^{(i)}
\Bigr) \\
& + \beta_{\text{KL}}\, D_{\text{KL}}\!\bigl(\pi_\theta \,\|\, \pi_{\text{ref}}\bigr),
\end{aligned}
\normalsize
\end{equation}
where $T_i$ is the length of trajectory $i$, $\mathcal{S}_{\text{val}}^{(i)}$ is the set of token positions strictly within valid structural boundaries, $\rho_t^{(i)}=\pi_\theta(y_t^{(i)}\mid y_{<t}^{(i)})/\pi_{\theta_{\text{old}}}(y_t^{(i)}\mid y_{<t}^{(i)})$ is the importance sampling ratio, $\epsilon_c$ is the PPO clipping range, $\beta_{\text{KL}}\geq 0$ controls the KL regularization strength, and $\pi_{\text{ref}}$ is a frozen reference policy. Unlike standard GRPO, SRA allows the gradient sign to vary within a single trajectory, enabling the model to reinforce desirable local behavior while penalizing undesirable local behavior within the same update.

\section{Experiments}
\label{sec:experiments}

We construct the experimental corpus from approximately 3,200 ACL conference papers published between 2021 and 2025. PDFs are parsed with MinerU~\citep{wang2024mineruopensourcesolutionprecise}, from which we extract titles, abstracts, introductions, figure/table captions, table contents, and available experimental metadata. ACL 2025 papers are held out exclusively for testing, yielding 1,176 test instances; the remaining ACL 2021--2024 papers are split into approximately 300 SFT papers, 1,550 RL-training papers, and 150 validation papers. Stage-level supervision is obtained by using GPT-4o~\citep{achiam2023gpt} with a fixed decomposition prompt to segment each introduction into background, problem, method, and contribution sections, together with corresponding outlines. For out-of-domain evaluation, we additionally collect 141 unseen CVPR papers, which are used only for zero-shot transfer evaluation and never for training or validation.

\begin{table}[t]
\centering
\setlength{\tabcolsep}{4pt} 
\begin{tabular}{lcccc}
\toprule
\textbf{Method} & \textbf{Sem.} & \textbf{Sec.} & \textbf{Len.} & \textbf{Struc.} \\
\midrule
\multicolumn{5}{c}{\textit{Qwen2.5-7B-Instruct Backbone}} \\
\midrule
Pure Prompt & 0.861 & 0.672 & 0.566 & 0.610 \\
Elaborate Prompt & 0.813 & 0.742 & 0.442 & 0.725 \\
AutoSurvey & 0.875 & 0.743 & 0.523 & 0.647 \\
STIG & 0.858 & 0.769 & 0.573 & 0.884 \\
Refinement w/o training & 0.878 & 0.746 & 0.693 & 0.800 \\
SurveyForge & 0.872 & 0.745 & 0.670 & 0.707 \\
\textbf{StructPO} & \textbf{0.913} & \textbf{0.825} & \textbf{0.708} & \textbf{0.971} \\
\midrule
\multicolumn{5}{c}{\textit{Qwen3-8B Backbone}} \\
\midrule
Pure Prompt & 0.871 & 0.773 & 0.666 & 0.761 \\
Elaborate Prompt & 0.862 & 0.741 & \textbf{0.703} & 0.802 \\
AutoSurvey & 0.884 & 0.749 & 0.609 & 0.727 \\
STIG & 0.870 & 0.789 & 0.487 & 0.840 \\
Refinement w/o training & 0.897 & 0.753 & 0.603 & 0.808 \\
SurveyForge & 0.878 & 0.721 & 0.547 & 0.755 \\
\textbf{StructPO} & \textbf{0.915} & \textbf{0.825} & 0.699 & \textbf{0.987} \\
\bottomrule
\end{tabular}%
\caption{Quantitative results of introduction generation on the ACL 2025 dataset. We compare StructPO against open-source prompt-based, workflow-based, and refinement baselines on automated metrics. Sem.: Semantic Similarity; Sec.: Section Similarity; Len.: Length Score; Struc.: Structure Score.}
\label{tab:main_results}
\end{table}

\begin{table*}[t]
\centering
\setlength{\tabcolsep}{3pt}
\renewcommand{\arraystretch}{1.15}
\fontsize{10pt}{12pt}\selectfont
\begin{tabular}{l cccc c ccccc c}
\toprule
\textbf{Method} 
& \textbf{Sem.} & \textbf{Sec.} & \textbf{Len.} & \textbf{Struc.} 
& \textbf{Auto Overall}
& \textbf{AWQ} & \textbf{SFC} & \textbf{Sound.} & \textbf{Pre.} & \textbf{Con.} 
& \textbf{Judge Overall} \\
\midrule
StructPO (Qwen2.5-7B) 
& 0.913 & 0.825 & 0.708 & 0.971 & 3.417 
& 3.095 & 4.439 & 2.553 & 2.981 & 2.874 & 3.188 \\
StructPO (Qwen3-8B) 
& 0.915 & 0.825 & 0.699 & \textbf{0.987} & \textbf{3.426} 
& 3.491 & 4.629 & 2.757 & 3.437 & 3.126 & 3.488 \\
StructPO (Qwen3-32B) 
& 0.914 & 0.825 & 0.701 & 0.982 & 3.422 
& 4.098 & 4.934 & 3.178 & 3.533 & 3.346 & 3.818 \\
\midrule
GPT-4o 
& 0.866 & 0.625 & \textbf{0.745} & 0.888 & 3.124 
& 2.959 & \textbf{4.993} & 2.951 & 3.039 & 2.913 & 3.371 \\
GPT-5.1 
& 0.915 & 0.822 & 0.656 & 0.941 & 3.334 
& 4.063 & 4.970 & 3.187 & 3.317 & \textbf{3.697} & \textbf{3.847} \\
GPT-5.5 
& 0.908 & 0.807 & 0.480 & 0.877 & 3.072 
& \textbf{4.199} & 4.992 & 3.117 & 3.077 & 3.123 & 3.702 \\
Claude Opus 4.8 
& \textbf{0.922} & \textbf{0.828} & 0.660 & 0.958 & 3.368 
& 4.073 & 4.983 & \textbf{3.429} & 3.394 & 3.291 & 3.834 \\
Gemini-3.1-Pro-Preview 
& 0.890 & 0.787 & 0.703 & 0.960 & 3.340 
& 4.108 & 4.983 & 3.233 & \textbf{3.618} & 3.020 & 3.792 \\
\bottomrule
\end{tabular}
\caption{Comprehensive evaluation of StructPO and strong closed-source LLMs.
Auto Overall is the sum of four automated metrics: Sem., Sec., Len., and Struc. Judge Overall is the average of five LLM-as-a-judge metrics: AWQ, SFC, Soundness, Presentation, and Contribution. Soundness, Presentation, and Contribution are reviewer-style rubric metrics used only for evaluation and are not used during StructPO training.}
\label{tab:main_results_qual}
\end{table*}

\subsection{Baselines and Models}
We instantiate StructPO on three open-source backbones of increasing scale: Qwen2.5-7B-Instruct, Qwen3-8B, and Qwen3-32B \citep{qwen2,yang2025qwen3technicalreport}, using Verl as the training framework and SGLang for rollout and inference \citep{sheng2025hybridflow,zheng2024sglang}. The 7B and 8B variants are trained on 8 H800 GPUs with a global batch size of 16 and a rollout size of 8; the 32B variant follows the same training recipe with adjusted parallelism to fit the larger model. We compare against Pure Prompt, ELABORATE Prompting \citep{garg2025let}, AutoSurvey \citep{NEURIPS2024_d07a9fc7}, STIG \citep{zhang2025eliminating}, SurveyForge \citep{DBLP:journals/corr/abs-2503-04629}, refinement without training and a suite of strong closed-source LLMs including GPT-4o, GPT-5.1, GPT-5.5, Claude Opus 4.8 and Gemini-3.1-Pro-Preview. All comparable open-source baselines use the same test set, input metadata, and decoding protocol.

\subsection{Evaluation Metrics}
We evaluate generated introductions with automated and LLM-as-a-judge metrics. Automated metrics include full-introduction semantic similarity (\textbf{Sem.}), section-level similarity (\textbf{Sec.}), length score (\textbf{Len.}), structure score (\textbf{Struc.}), and \textbf{Auto Overall}, the sum of the four normalized scores. Following \citet{zheng2023judging}, judge metrics include Academic Writing Quality (\textbf{AWQ}), Scientific and Factual Consistency (\textbf{SFC}), \textbf{Soundness}, \textbf{Presentation}, and \textbf{Contribution}; their average is reported as \textbf{Judge Overall}. Soundness assesses accurate and efficient reflection of core paper information, Presentation assesses rhetorical placement and coherent flow, and Contribution assesses whether the main contributions are stated clearly, specifically, and concisely. 

These three reviewer-style metrics are used only for evaluation, not for StructPO training. For the main comparison in Table~\ref{tab:main_results_qual} we report all five judge metrics, while the ablation in Table~\ref{tab:ablation_results} reports only AWQ and SFC to keep the table compact, as these two metrics most directly capture writing quality and factual grounding. All methods are evaluated with the same fixed judge prompt and scoring rubric.

\subsection{Main Results}

Table~\ref{tab:main_results} shows that StructPO consistently improves semantic alignment, section-level quality, and structural rationality over prompt- and workflow-based baselines on both open-source backbones. Although AutoSurvey decomposes writing into stages, its external workflow still underperforms a single-pass policy trained with stage-level credit assignment, suggesting that decomposition must be internalized to avoid cross-stage drift. STIG obtains strong structure scores through stage-token generation, but suffers from length collapse. In contrast, StructPO preserves structural control while producing richer and more semantically aligned content.

\begin{table*}[t]
\centering
\begin{tabular}{lccccccc}
\toprule
\textbf{Method} & \textbf{Sem.} ($\uparrow$) & \textbf{Sec.} ($\uparrow$) & \textbf{Len.} ($\uparrow$) & \textbf{Struc.} ($\uparrow$) & \textbf{AWQ} ($\uparrow$) & \textbf{SFC} ($\uparrow$) & \textbf{Overall} ($\uparrow$) \\
\midrule
\multicolumn{8}{c}{\textit{Base Model: Qwen2.5-7B-Instruct}} \\
\midrule
Refinement w/o training & 0.878 & 0.746 & 0.693 & 0.800 & 2.652 & 3.290 & 74.214 \\
SFT & 0.885 & 0.786 & 0.547 & 0.832 & 2.929 & 4.419 & 75.682 \\
GRPO & \textbf{0.921} & \textbf{0.829} & 0.694 & \textbf{0.986} & 3.080 & 2.258 & 79.290 \\
StructPO w/o Refinement & 0.912 & 0.828 & 0.697 & 0.959 & 3.060 & 4.366 & 82.764 \\
StructPO w/o Global Reward & 0.910 & 0.821 & 0.706 & 0.975 & 3.009 & 4.357 & 82.976 \\
StructPO w/o Struct-Awareness & 0.908 & 0.823 & 0.644 & 0.985 & \textbf{3.262} & \textbf{4.502} & 82.734 \\
\textbf{StructPO (Ours)} & 0.913 & 0.825 & \textbf{0.708} & 0.971 & 3.095 & 4.439 & \textbf{83.398} \\
\midrule
\multicolumn{8}{c}{\textit{Base Model: Qwen3-8B}} \\
\midrule
Refinement w/o training & 0.897 & 0.753 & 0.603 & 0.808 & 2.772 & \textbf{4.974} & 76.674 \\
SFT & 0.886 & 0.790 & 0.529 & 0.830 & 3.021 & 4.557 & 75.849 \\
GRPO & 0.904 & 0.788 & 0.663 & 0.984 & 3.337 & 4.972 & 83.387 \\
StructPO w/o Refinement & \textbf{0.915} & \textbf{0.827} & 0.663 & 0.983 & 3.504 & 4.724 & 84.217 \\
StructPO w/o Global Reward & 0.908 & 0.821 & 0.638 & 0.981 & \textbf{3.606} & 4.884 & 83.932 \\
StructPO w/o Struct-Awareness & 0.904 & 0.820 & 0.523 & 0.955 & 3.421 & 4.856 & 80.590 \\
\textbf{StructPO (Ours)} & \textbf{0.915} & 0.825 & \textbf{0.699} & \textbf{0.987} & 3.491 & 4.629 & \textbf{84.734} \\
\bottomrule
\end{tabular}
\caption{Ablation study on different components of StructPO. Refinement w/o training: Inference-time refinement using SFT model; SFT: Supervised Fine-Tuning baseline; GRPO: Standard trajectory-level policy optimization; w/o Refinement: Removed refinement-guided penalty; w/o Global: Removed global semantic alignment reward; w/o Struct-Aware: Removed stage-level local advantage assignment. For brevity, we report AWQ and SFC as representative judge metrics; Overall is a composite score defined as the sum of the four automated metrics (rescaled to $[0,100]$) and the two judge metrics, so its scale differs from the Auto Overall in Table~\ref{tab:main_results_qual}. The best results are highlighted in bold.}
\label{tab:ablation_results}
\end{table*}

\begin{figure*}[t]
  \centering
  \includegraphics[width=0.85\textwidth]{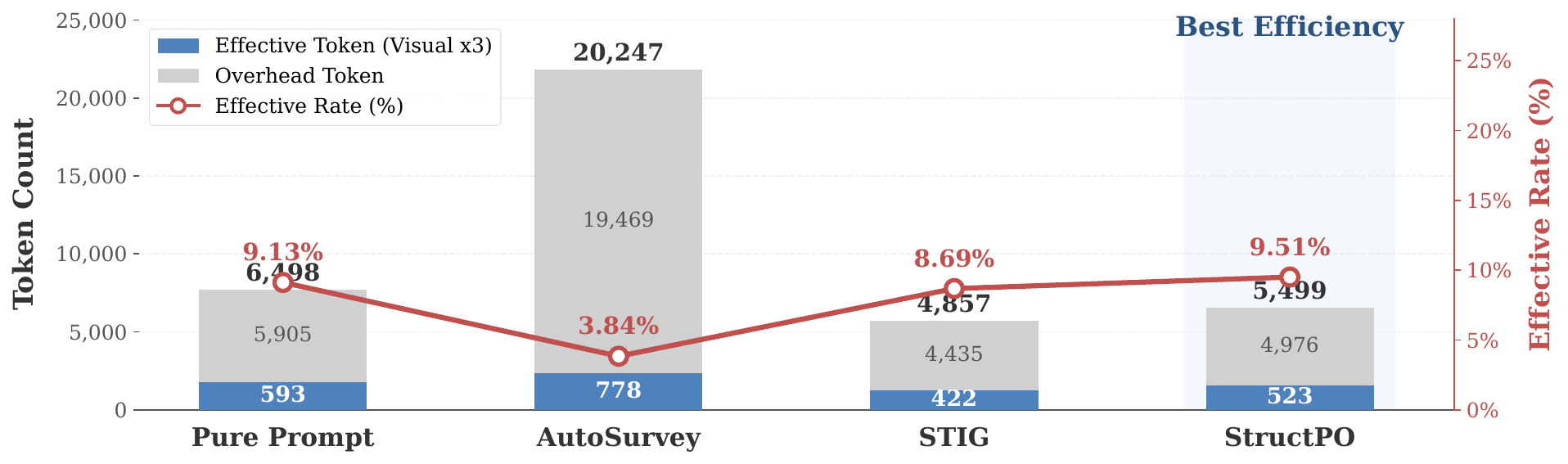}
  \caption{Inference-token comparison across generation paradigms. StructPO achieves a favorable Pareto balance between informative content and computational overhead, avoiding both the token bloat of AutoSurvey and the length collapse of STIG. Effective token bars are visually scaled 3$\times$ for clarity.}
  \label{fig:token_comparison}
\end{figure*}

\begin{figure*}[ht]
  \centering
  \includegraphics[width=0.9\textwidth]{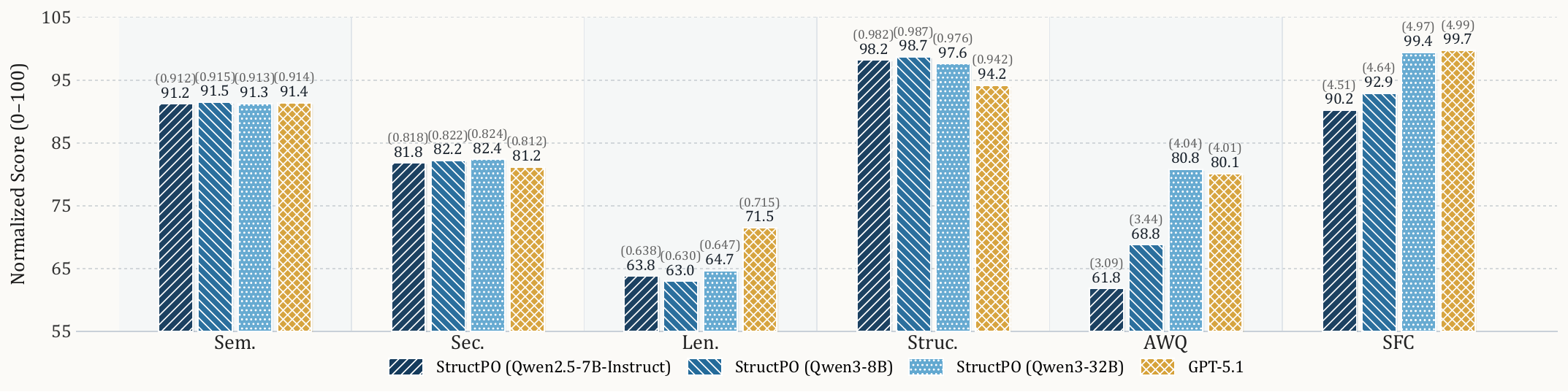}
  \caption{Zero-shot transfer results from ACL to 141 unseen CVPR papers. StructPO variants preserve strong structural control under domain shift, consistently outperforming GPT-5.1 on structure score and section-level alignment. GPT-5.1 retains an advantage in length control and factual consistency, while StructPO with Qwen3-32B achieves the best academic writing quality score.}
  \label{fig:generalization_study}
\end{figure*}

Table~\ref{tab:main_results_qual} compares StructPO with strong closed-source LLMs. On automated metrics, all StructPO variants achieve higher Auto Overall scores than the closed-source models, mainly through stronger section alignment and structural rationality. On judge metrics, closed-source LLMs retain advantages in factual reliability and contribution framing, likely due to stronger pretrained knowledge and instruction following. However, scaling StructPO from 8B to 32B substantially improves Judge Overall, surpassing GPT-5.5 and Gemini-3.1-Pro-Preview and nearly matching GPT-5.1 and Claude Opus 4.8. These results suggest that StructPO provides robust structural control, while larger backbones further improve content quality. These findings suggest that StructPO is complementary to backbone scaling, with policy optimization providing reliable rhetorical structure and larger models improving factual grounding and fluency.

\section{Analysis}
\label{sec:analysis}

\subsection{Ablation Study}

Table~\ref{tab:ablation_results} shows the contribution of each component. SFT learns the stage-token format but suffers from length collapse, especially on Qwen3-8B where the length score is only 0.529, indicating that supervised imitation captures surface structure but not long-form rhetorical completeness.

Standard GRPO improves several automatic metrics but is less stable because it assigns one trajectory-level advantage to all tokens; on Qwen2.5-7B, this causes a large drop in factual and citation consistency. Removing struct-aware advantage assignment also degrades performance, particularly length control on Qwen3-8B, suggesting that heterogeneous stages require separate credit assignment. The global reward and refinement-guided penalty provide complementary gains: without the global reward, semantic alignment weakens; without refinement guidance, the model loses part of the benefit from draft--revision comparisons. StructPO achieves the best overall performance on both backbones, confirming that local stage rewards, global alignment, and refinement-guided training jointly improve structured introduction generation. The ablation results also indicate that no single component dominates across all metrics; instead, StructPO benefits from balancing local controllability, global semantic alignment, and revision-aware regularization. We omit full Qwen3-32B ablations due to training cost, but trends across 7B and 8B suggest similar conclusions at larger scale.

\subsection{Qualitative Case Study}

We examine a representative ACL 2025 test case, the SpeechFake paper~\citep{huang-etal-2025-speechfake}. AutoSurvey generates detailed content but exhibits cross-stage drift, such as revealing the proposed dataset too early and producing markdown-style meta-commentary. STIG preserves the four-part skeleton, but its output is short, repetitive, and weak in contribution synthesis.

StructPO produces a more balanced introduction. It follows the background--problem--method--contribution progression with clearer transitions, while incorporating key details such as dataset characteristics, equal error rate, and cross-speaker evaluation. Compared with AutoSurvey, it avoids workflow fragmentation; compared with STIG, it provides denser content and stronger contribution framing. This qualitative pattern is consistent with the quantitative results, where StructPO improves structure without collapsing output length. A full side-by-side example is included in the Appendix \ref{app:case}.

\subsection{Token Efficiency Analysis}

We compare inference-time token consumption using Qwen3-8B as the common backbone. As shown in Figure~\ref{fig:token_comparison}, total usage is divided into \textit{overhead tokens} (input context, prompts, and intermediate workflow states) and \textit{effective tokens} (final introduction content), with the effective rate defined as the ratio of effective tokens to total tokens. Workflow-based systems incur large overhead due to repeated context replay: AutoSurvey generates 778 effective tokens but requires 19,469 overhead tokens, yielding only a 3.84\% effective rate. In contrast, single-pass baselines reduce orchestration cost but may sacrifice completeness; STIG produces only 422 effective tokens, indicating length collapse.

StructPO achieves a better efficiency--completeness trade-off. Since refinement behavior is distilled into the policy during training, StructPO performs single-pass inference without external revision loops. It generates 523 effective tokens with 4,976 overhead tokens, reducing overhead by 74.4\% compared with AutoSurvey while producing richer content than STIG. This shows that internalizing writing workflows can substantially improve deployment efficiency without costly multi-turn orchestration.

\subsection{Generalization Study}

To assess cross-domain generalization, we evaluate ACL-trained checkpoints on 141 unseen CVPR papers without domain-specific tuning. As shown in Figure~\ref{fig:generalization_study}, StructPO transfers its rhetorical scaffolding effectively: all variants preserve strong structural rationality and outperform GPT-5.1 on structure and section-level alignment, despite GPT-5.1's broader exposure to vision-domain text.

On content-oriented judge metrics, StructPO-32B is close to GPT-5.1, matching factual consistency and slightly improving academic writing quality. Semantic similarity is also comparable across systems. These results suggest that stage-level rewards learn a domain-general introduction structure with strong transferability, while remaining weaknesses under domain shift are mainly stylistic and length-related. Adaptive length targets and retrieval augmentation may further improve transfer.

\subsection{Human Evaluation}

We conduct a blind human evaluation on 30 randomly sampled ACL test papers, comparing StructPO with Qwen3-32B against GPT-5.1. Three NLP researchers judge anonymized outputs based on coherence, structural completeness and academic writing quality. As shown in Table~\ref{tab:human_eval}, StructPO achieves a 53.3\% win rate by majority voting and receives 55.6\% of all annotator votes. These results indicate that, with a stronger open-source backbone, StructPO can be preferred over GPT-5.1 in human evaluation, although the margin remains moderate.

\begin{table}[t]
\centering
\begin{tabular}{lcc}
\toprule
\textbf{Comparison} & \textbf{StructPO Win} & \textbf{GPT-5.1 Win} \\
\midrule
Win Rate & 53.3\% (16/30) & 46.7\% (14/30) \\
Win Votes & 55.6\% (50/90) & 44.4\% (40/90) \\
\bottomrule

\end{tabular}
\caption{Human evaluation results comparing StructPO with Qwen3-32B against GPT-5.1. Win Rate is computed by majority voting over 30 samples, and Win Votes are computed from all 90 annotator votes.}
\label{tab:human_eval}
\end{table}

\section{Conclusion}
\label{sec:conclusion}

We proposed StructPO, a structure-aware policy optimization framework that internalizes multi-stage academic introduction writing into a single-pass stage-token policy. By combining stage-level credit assignment with refinement-guided optimization, StructPO improves local rhetorical quality, global coherence, and inference efficiency. Experiments on ACL papers show consistent gains over prompt- and workflow-based baselines, and zero-shot evaluation on CVPR papers demonstrates effective transfer. StructPO still relies on a fixed eight-stage template and may require adaptation for theoretical, survey, or non-standard papers. Future work may explore adaptive structural planning and retrieval-augmented grounding within the SRA reward loop \citep{lewis2020retrieval}.

\bibliography{aaai2027}

@article{zhang2025exploring,
  title={Exploring the role of large language models in the scientific method: from hypothesis to discovery},
  author={Zhang, Yanbo and Khan, Sumeer A and Mahmud, Adnan and Yang, Huck and Lavin, Alexander and Levin, Michael and Frey, Jeremy and Dunnmon, Jared and Evans, James and Bundy, Alan and others},
  journal={npj Artificial Intelligence},
  volume={1},
  number={1},
  pages={14},
  year={2025},
  publisher={Nature Publishing Group UK London}
}

@inproceedings{garikaparthi-etal-2025-mir,
    title = "{MIR}: Methodology Inspiration Retrieval for Scientific Research Problems",
    author = "Garikaparthi, Aniketh  and
      Patwardhan, Manasi  and
      Kanade, Aditya Sanjiv  and
      Hassan, Aman  and
      Vig, Lovekesh  and
      Cohan, Arman",
    editor = "Che, Wanxiang  and
      Nabende, Joyce  and
      Shutova, Ekaterina  and
      Pilehvar, Mohammad Taher",
    booktitle = "Proceedings of the 63rd Annual Meeting of the Association for Computational Linguistics (Volume 1: Long Papers)",
    month = jul,
    year = "2025",
    address = "Vienna, Austria",
    publisher = "Association for Computational Linguistics",
    url = "https://aclanthology.org/2025.acl-long.1390/",
    doi = "10.18653/v1/2025.acl-long.1390",
    pages = "28614--28659",
    ISBN = "979-8-89176-251-0"
}

@inproceedings{
li2025scilitllm,
title={SciLit{LLM}: How to Adapt {LLM}s for Scientific Literature Understanding},
author={Sihang Li and Jin Huang and Jiaxi Zhuang and Yaorui Shi and Xiaochen Cai and Mingjun Xu and Xiang Wang and Linfeng Zhang and Guolin Ke and Hengxing Cai},
booktitle={The Thirteenth International Conference on Learning Representations},
year={2025},
url={https://openreview.net/forum?id=8dzKkeWUUb}
}

@inproceedings{yang-etal-2024-large-language,
    title = "Large Language Models for Automated Open-domain Scientific Hypotheses Discovery",
    author = "Yang, Zonglin  and
      Du, Xinya  and
      Li, Junxian  and
      Zheng, Jie  and
      Poria, Soujanya  and
      Cambria, Erik",
    editor = "Ku, Lun-Wei  and
      Martins, Andre  and
      Srikumar, Vivek",
    booktitle = "Findings of the Association for Computational Linguistics: ACL 2024",
    month = aug,
    year = "2024",
    address = "Bangkok, Thailand",
    publisher = "Association for Computational Linguistics",
    url = "https://aclanthology.org/2024.findings-acl.804/",
    doi = "10.18653/v1/2024.findings-acl.804",
    pages = "13545--13565"
}

@article{boiko2023autonomous,
  title={Autonomous chemical research with large language models},
  author={Boiko, Daniil A and MacKnight, Robert and Kline, Ben and Gomes, Gabe},
  journal={Nature},
  volume={624},
  number={7992},
  pages={570--578},
  year={2023},
  publisher={Nature Publishing Group UK London}
}

@inproceedings{zhao-etal-2025-abgen,
    title = "{A}b{G}en: Evaluating Large Language Models in Ablation Study Design and Evaluation for Scientific Research",
    author = "Zhao, Yilun  and
      Chen, Weiyuan  and
      Xu, Zhijian  and
      Patwardhan, Manasi  and
      Wang, Chengye  and
      Liu, Yixin  and
      Vig, Lovekesh  and
      Cohan, Arman",
    editor = "Che, Wanxiang  and
      Nabende, Joyce  and
      Shutova, Ekaterina  and
      Pilehvar, Mohammad Taher",
    booktitle = "Proceedings of the 63rd Annual Meeting of the Association for Computational Linguistics (Volume 1: Long Papers)",
    month = jul,
    year = "2025",
    address = "Vienna, Austria",
    publisher = "Association for Computational Linguistics",
    url = "https://aclanthology.org/2025.acl-long.611/",
    doi = "10.18653/v1/2025.acl-long.611",
    pages = "12479--12491",
    ISBN = "979-8-89176-251-0"
}

@article{DBLP:journals/corr/abs-2504-18765,
  publtype={informal},
  author={Chengwei Liu and Chong Wang and Jiayue Cao and Jingquan Ge and Kun Wang and Lvye Zhang and Ming-Ming Cheng and Penghai Zhao and Tianlin Li and Xiaojun Jia and Xiang Li and Xinfeng Li and Yang Liu and Yebo Feng and Yihao Huang and Yijia Xu and Yuqiang Sun and Zhenhong Zhou and Zhengzi Xu},
  title={A Vision for Auto Research with LLM Agents},
  year={2025},
  month={April},
  cdate={1743465600000},
  journal={CoRR},
  volume={abs/2504.18765},
  url={https://doi.org/10.48550/arXiv.2504.18765}
}

@inproceedings{NEURIPS2024_d07a9fc7,
 author = {Wang, Yidong and Guo, Qi and Yao, Wenjin and Zhang, Hongbo and Zhang, Xin and Wu, Zhen and Zhang, Meishan and Dai, Xinyu and Zhang, Min and Wen, Qingsong and Ye, Wei and Zhang, Shikun and Zhang, Yue},
 booktitle = {Advances in Neural Information Processing Systems},
 editor = {A. Globerson and L. Mackey and D. Belgrave and A. Fan and U. Paquet and J. Tomczak and C. Zhang},
 pages = {115119--115145},
 publisher = {Curran Associates, Inc.},
 title = {AutoSurvey: Large Language Models Can Automatically Write Surveys},
 url = {https://proceedings.neurips.cc/paper_files/paper/2024/file/d07a9fc7da2e2ec0574c38d5f504d105-Paper-Conference.pdf},
 volume = {37},
 year = {2024}
}

@article{DBLP:journals/corr/abs-2503-04629,
  publtype={informal},
  author={Xiangchao Yan and Shiyang Feng and Jiakang Yuan and Renqiu Xia and Bin Wang and Bo Zhang and Lei Bai},
  title={SurveyForge: On the Outline Heuristics, Memory-Driven Generation, and Multi-dimensional Evaluation for Automated Survey Writing},
  year={2025},
  month={March},
  cdate={1740787200000},
  journal={CoRR},
  volume={abs/2503.04629},
  url={https://doi.org/10.48550/arXiv.2503.04629}
}

@inproceedings{wang2021autocite,
  title={Autocite: Multi-modal representation fusion for contextual citation generation},
  author={Wang, Qingqin and Xiong, Yun and Zhang, Yao and Zhang, Jiawei and Zhu, Yangyong},
  booktitle={Proceedings of the 14th ACM International Conference on Web Search and Data Mining},
  pages={788--796},
  year={2021}
}

@inproceedings{ge-etal-2021-baco,
    title = "{BACO}: A Background Knowledge- and Content-Based Framework for Citing Sentence Generation",
    author = "Ge, Yubin  and
      Dinh, Ly  and
      Liu, Xiaofeng  and
      Su, Jinsong  and
      Lu, Ziyao  and
      Wang, Ante  and
      Diesner, Jana",
    editor = "Zong, Chengqing  and
      Xia, Fei  and
      Li, Wenjie  and
      Navigli, Roberto",
    booktitle = "Proceedings of the 59th Annual Meeting of the Association for Computational Linguistics and the 11th International Joint Conference on Natural Language Processing (Volume 1: Long Papers)",
    month = aug,
    year = "2021",
    address = "Online",
    publisher = "Association for Computational Linguistics",
    url = "https://aclanthology.org/2021.acl-long.116/",
    doi = "10.18653/v1/2021.acl-long.116",
    pages = "1466--1478"
}

@article{garg2025let,
  title={Let's Use ChatGPT To Write Our Paper! Benchmarking LLMs To Write the Introduction of a Research Paper},
  author={Garg, Krishna and Shaikh, Firoz and Bandyopadhyay, Sambaran and Caragea, Cornelia},
  journal={arXiv preprint arXiv:2508.14273},
  year={2025}
}

@misc{schmidgall2025agentlaboratoryusingllm,
      title={Agent Laboratory: Using LLM Agents as Research Assistants}, 
      author={Samuel Schmidgall and Yusheng Su and Ze Wang and Ximeng Sun and Jialian Wu and Xiaodong Yu and Jiang Liu and Michael Moor and Zicheng Liu and Emad Barsoum},
      year={2025},
      eprint={2501.04227},
      archivePrefix={arXiv},
      primaryClass={cs.HC},
      url={https://arxiv.org/abs/2501.04227}, 
}

@inproceedings{su-etal-2025-many,
    title = "Many Heads Are Better Than One: Improved Scientific Idea Generation by A {LLM}-Based Multi-Agent System",
    author = "Su, Haoyang  and
      Chen, Renqi  and
      Tang, Shixiang  and
      Yin, Zhenfei  and
      Zheng, Xinzhe  and
      Li, Jinzhe  and
      Qi, Biqing  and
      Wu, Qi  and
      Li, Hui  and
      Ouyang, Wanli  and
      Torr, Philip  and
      Zhou, Bowen  and
      Dong, Nanqing",
    editor = "Che, Wanxiang  and
      Nabende, Joyce  and
      Shutova, Ekaterina  and
      Pilehvar, Mohammad Taher",
    booktitle = "Proceedings of the 63rd Annual Meeting of the Association for Computational Linguistics (Volume 1: Long Papers)",
    month = jul,
    year = "2025",
    address = "Vienna, Austria",
    publisher = "Association for Computational Linguistics",
    url = "https://aclanthology.org/2025.acl-long.1368/",
    doi = "10.18653/v1/2025.acl-long.1368",
    pages = "28201--28240",
    ISBN = "979-8-89176-251-0"
}

@misc{wang2024mineruopensourcesolutionprecise,
      title={MinerU: An Open-Source Solution for Precise Document Content Extraction}, 
      author={Bin Wang and Chao Xu and Xiaomeng Zhao and Linke Ouyang and Fan Wu and Zhiyuan Zhao and Rui Xu and Kaiwen Liu and Yuan Qu and Fukai Shang and Bo Zhang and Liqun Wei and Zhihao Sui and Wei Li and Botian Shi and Yu Qiao and Dahua Lin and Conghui He},
      year={2024},
      eprint={2409.18839},
      archivePrefix={arXiv},
      primaryClass={cs.CV},
      url={https://arxiv.org/abs/2409.18839}, 
}

@article{qwen2,
      title={Qwen2 Technical Report}, 
      author={An Yang and Baosong Yang and Binyuan Hui and Bo Zheng and Bowen Yu and Chang Zhou and Chengpeng Li and Chengyuan Li and Dayiheng Liu and Fei Huang and Guanting Dong and Haoran Wei and Huan Lin and Jialong Tang and Jialin Wang and Jian Yang and Jianhong Tu and Jianwei Zhang and Jianxin Ma and Jin Xu and Jingren Zhou and Jinze Bai and Jinzheng He and Junyang Lin and Kai Dang and Keming Lu and Keqin Chen and Kexin Yang and Mei Li and Mingfeng Xue and Na Ni and Pei Zhang and Peng Wang and Ru Peng and Rui Men and Ruize Gao and Runji Lin and Shijie Wang and Shuai Bai and Sinan Tan and Tianhang Zhu and Tianhao Li and Tianyu Liu and Wenbin Ge and Xiaodong Deng and Xiaohuan Zhou and Xingzhang Ren and Xinyu Zhang and Xipin Wei and Xuancheng Ren and Yang Fan and Yang Yao and Yichang Zhang and Yu Wan and Yunfei Chu and Yuqiong Liu and Zeyu Cui and Zhenru Zhang and Zhihao Fan},
      journal={arXiv preprint arXiv:2407.10671},
      year={2024}
}

@article{achiam2023gpt,
  title={Gpt-4 technical report},
  author={Achiam, Josh and Adler, Steven and Agarwal, Sandhini and Ahmad, Lama and Akkaya, Ilge and Aleman, Florencia Leoni and Almeida, Diogo and Altenschmidt, Janko and Altman, Sam and Anadkat, Shyamal and others},
  journal={arXiv preprint arXiv:2303.08774},
  year={2023}
}

@article{wu2023autogen,
  title={Autogen: Enabling next-gen llm applications via multi-agent conversation framework},
  author={Wu, Qingyun and Bansal, Gagan and Zhang, Jieyu and Wu, Yiran and Zhang, Shaokun and Zhu, Erkang and Li, Beibin and Jiang, Li and Zhang, Xiaoyun and Wang, Chi},
  journal={arXiv preprint arXiv:2308.08155},
  volume={3},
  number={4},
  year={2023}
}

@article{DBLP:journals/corr/abs-2408-06292,
  publtype={informal},
  author={Chris Lu and Cong Lu and Robert Tjarko Lange and Jakob N. Foerster and Jeff Clune and David Ha},
  title={The AI Scientist: Towards Fully Automated Open-Ended Scientific Discovery},
  year={2024},
  cdate={1704067200000},
  journal={CoRR},
  volume={abs/2408.06292},
  url={https://doi.org/10.48550/arXiv.2408.06292}
}

@inproceedings{baek-etal-2025-researchagent,
    title = "{R}esearch{A}gent: Iterative Research Idea Generation over Scientific Literature with Large Language Models",
    author = "Baek, Jinheon  and
      Jauhar, Sujay Kumar  and
      Cucerzan, Silviu  and
      Hwang, Sung Ju",
    editor = "Chiruzzo, Luis  and
      Ritter, Alan  and
      Wang, Lu",
    booktitle = "Proceedings of the 2025 Conference of the Nations of the Americas Chapter of the Association for Computational Linguistics: Human Language Technologies (Volume 1: Long Papers)",
    month = apr,
    year = "2025",
    address = "Albuquerque, New Mexico",
    publisher = "Association for Computational Linguistics",
    url = "https://aclanthology.org/2025.naacl-long.342/",
    doi = "10.18653/v1/2025.naacl-long.342",
    pages = "6709--6738",
    ISBN = "979-8-89176-189-6"
}

@article{wang2023scientific,
  title={Scientific discovery in the age of artificial intelligence},
  author={Wang, Hanchen and Fu, Tianfan and Du, Yuanqi and Gao, Wenhao and Huang, Kexin and Liu, Ziming and Chandak, Payal and Liu, Shengchao and Van Katwyk, Peter and Deac, Andreea and others},
  journal={Nature},
  volume={620},
  number={7972},
  pages={47--60},
  year={2023},
  publisher={Nature Publishing Group UK London}
}

@article{Langley_2024, title={Integrated Systems for Computational Scientific Discovery}, volume={38}, url={https://ojs.aaai.org/index.php/AAAI/article/view/30269}, DOI={10.1609/aaai.v38i20.30269}, abstractNote={This paper poses the challenge of developing and evaluating integrated
systems for computational scientific discovery. We note some distinguishing
characteristics of discovery tasks, examine eight component abilities,
review previous successes at partial integration, and consider hurdles
the AI research community must leap to transform the vision for
integrated discovery into reality. In closing, we discuss promising
scientific domains in which to test such computational artifacts.}, number={20}, journal={Proceedings of the AAAI Conference on Artificial Intelligence}, author={Langley, Pat}, year={2024}, month={Mar.}, pages={22598-22606} }

@inproceedings{zimmermann2024leveraging,
  title={Leveraging large language models for literature review tasks-a case study using chatgpt},
  author={Zimmermann, Robert and Staab, Marina and Nasseri, Mehran and Brandtner, Patrick},
  booktitle={International Conference on Advanced Research in Technologies, Information, Innovation and Sustainability},
  pages={313--323},
  year={2024},
  organization={Springer}
}

@article{agarwal2024litllm,
  title={Litllm: A toolkit for scientific literature review},
  author={Agarwal, Shubham and Sahu, Gaurav and Puri, Abhay and Laradji, Issam H and Dvijotham, Krishnamurthy DJ and Stanley, Jason and Charlin, Laurent and Pal, Christopher},
  journal={arXiv preprint arXiv:2402.01788},
  year={2024}
}

@misc{zhang2025eliminating,
      title={Eliminating Agentic Workflow for Introduction Generation with Parametric Stage Tokens}, 
      author={Meicong Zhang and Tiancheng su and Guoxiu He},
      year={2025},
      eprint={2601.09728},
      archivePrefix={arXiv},
      primaryClass={cs.CL},
      url={https://arxiv.org/abs/2601.09728}, 
}

@misc{shao2024grpo,
      title={DeepSeekMath: Pushing the Limits of Mathematical Reasoning in Open Language Models}, 
      author={Zhihong Shao and Peiyi Wang and Qihao Zhu and Runxin Xu and Junxiao Song and Xiao Bi and Haowei Zhang and Mingchuan Zhang and Y. K. Li and Y. Wu and Daya Guo},
      year={2024},
      eprint={2402.03300},
      archivePrefix={arXiv},
      primaryClass={cs.CL},
      url={https://arxiv.org/abs/2402.03300}, 
}

@inproceedings{
feng2025groupingroup,
title={Group-in-Group Policy Optimization for {LLM} Agent Training},
author={Lang Feng and Zhenghai Xue and Tingcong Liu and Bo An},
booktitle={The Thirty-ninth Annual Conference on Neural Information Processing Systems},
year={2025},
url={https://openreview.net/forum?id=QXEhBMNrCW}
}

@inproceedings{
li2026ssvpo,
title={{SSVPO}: Effective Step-Level Credit Assignment for {RL} Training of Language Models},
author={Yugu Li and Zehong Cao and Jianglin Qiao and Siyi Hu},
booktitle={The Fourteenth International Conference on Learning Representations},
year={2026},
url={https://openreview.net/forum?id=g33DGvnHYd}
}

@inproceedings{
wang2026information,
title={Information Gain-based Policy Optimization: A Simple and Effective Approach for Multi-Turn Search Agents},
author={Guoqing Wang and Sunhao Dai and Guangze Ye and Zeyu Gan and Wei Yao and Yong Deng and Xiaofeng Wu and Zhenzhe Ying},
booktitle={The Fourteenth International Conference on Learning Representations},
year={2026},
url={https://openreview.net/forum?id=qkWP6phrvZ}
}

@misc{fei2025selfguidedprocessrewardoptimization,
      title={Self-Guided Process Reward Optimization with Redefined Step-wise Advantage for Process Reinforcement Learning}, 
      author={Wu Fei and Hao Kong and Shuxian Liang and Yang Lin and Yibo Yang and Jing Tang and Lei Chen and Xiansheng Hua},
      year={2025},
      eprint={2507.01551},
      archivePrefix={arXiv},
      primaryClass={cs.LG},
      url={https://arxiv.org/abs/2507.01551}, 
}

@inproceedings{sheng2025hybridflow,
  title={Hybridflow: A flexible and efficient rlhf framework},
  author={Sheng, Guangming and Zhang, Chi and Ye, Zilingfeng and Wu, Xibin and Zhang, Wang and Zhang, Ru and Peng, Yanghua and Lin, Haibin and Wu, Chuan},
  booktitle={Proceedings of the Twentieth European Conference on Computer Systems},
  pages={1279--1297},
  year={2025}
}

@article{lewis2020retrieval,
  title={Retrieval-augmented generation for knowledge-intensive nlp tasks},
  author={Lewis, Patrick and Perez, Ethan and Piktus, Aleksandra and Petroni, Fabio and Karpukhin, Vladimir and Goyal, Naman and K{\"u}ttler, Heinrich and Lewis, Mike and Yih, Wen-tau and Rockt{\"a}schel, Tim and others},
  journal={Advances in neural information processing systems},
  volume={33},
  pages={9459--9474},
  year={2020}
}

@misc{yang2025qwen3technicalreport,
      title={Qwen3 Technical Report}, 
      author={An Yang and Anfeng Li and Baosong Yang and Beichen Zhang and Binyuan Hui and Bo Zheng and Bowen Yu and Chang Gao and Chengen Huang and Chenxu Lv and Chujie Zheng and Dayiheng Liu and Fan Zhou and Fei Huang and Feng Hu and Hao Ge and Haoran Wei and Huan Lin and Jialong Tang and Jian Yang and Jianhong Tu and Jianwei Zhang and Jianxin Yang and Jiaxi Yang and Jing Zhou and Jingren Zhou and Junyang Lin and Kai Dang and Keqin Bao and Kexin Yang and Le Yu and Lianghao Deng and Mei Li and Mingfeng Xue and Mingze Li and Pei Zhang and Peng Wang and Qin Zhu and Rui Men and Ruize Gao and Shixuan Liu and Shuang Luo and Tianhao Li and Tianyi Tang and Wenbiao Yin and Xingzhang Ren and Xinyu Wang and Xinyu Zhang and Xuancheng Ren and Yang Fan and Yang Su and Yichang Zhang and Yinger Zhang and Yu Wan and Yuqiong Liu and Zekun Wang and Zeyu Cui and Zhenru Zhang and Zhipeng Zhou and Zihan Qiu},
      year={2025},
      eprint={2505.09388},
      archivePrefix={arXiv},
      primaryClass={cs.CL},
      url={https://arxiv.org/abs/2505.09388}, 
}

@inproceedings{huang-etal-2025-speechfake,
    title = "{S}peech{F}ake: A Large-Scale Multilingual Speech Deepfake Dataset Incorporating Cutting-Edge Generation Methods",
    author = "Huang, Wen  and
      Gu, Yanmei  and
      Wang, Zhiming  and
      Zhu, Huijia  and
      Qian, Yanmin",
    editor = "Che, Wanxiang  and
      Nabende, Joyce  and
      Shutova, Ekaterina  and
      Pilehvar, Mohammad Taher",
    booktitle = "Proceedings of the 63rd Annual Meeting of the Association for Computational Linguistics (Volume 1: Long Papers)",
    month = jul,
    year = "2025",
    address = "Vienna, Austria",
    publisher = "Association for Computational Linguistics",
    url = "https://aclanthology.org/2025.acl-long.493/",
    doi = "10.18653/v1/2025.acl-long.493",
    pages = "9985--9998",
    ISBN = "979-8-89176-251-0"
}

@inproceedings{li-etal-2025-chain-ideas,
    title = "Chain of Ideas: Revolutionizing Research Via Novel Idea Development with {LLM} Agents",
    author = "Li, Long  and
      Xu, Weiwen  and
      Guo, Jiayan  and
      Zhao, Ruochen  and
      Li, Xingxuan  and
      Yuan, Yuqian  and
      Zhang, Boqiang  and
      Jiang, Yuming  and
      Xin, Yifei  and
      Dang, Ronghao  and
      Rong, Yu  and
      Zhao, Deli  and
      Feng, Tian  and
      Bing, Lidong",
    editor = "Christodoulopoulos, Christos  and
      Chakraborty, Tanmoy  and
      Rose, Carolyn  and
      Peng, Violet",
    booktitle = "Findings of the Association for Computational Linguistics: EMNLP 2025",
    month = nov,
    year = "2025",
    address = "Suzhou, China",
    publisher = "Association for Computational Linguistics",
    url = "https://aclanthology.org/2025.findings-emnlp.477/",
    doi = "10.18653/v1/2025.findings-emnlp.477",
    pages = "8971--9004",
    ISBN = "979-8-89176-335-7"
}

@misc{huang2026sketchvl,
      title={SketchVL: Policy Optimization via Fine-Grained Credit Assignment for Chart Understanding and More}, 
      author={Muye Huang and Lingling Zhang and Yifei Li and Yaqiang Wu and Jun Liu},
      year={2026},
      eprint={2601.05688},
      archivePrefix={arXiv},
      primaryClass={cs.CV},
      url={https://arxiv.org/abs/2601.05688}, 
}

@misc{he2021debertav3,
      title={DeBERTaV3: Improving DeBERTa using ELECTRA-Style Pre-Training with Gradient-Disentangled Embedding Sharing}, 
      author={Pengcheng He and Jianfeng Gao and Weizhu Chen},
      year={2021},
      eprint={2111.09543},
      archivePrefix={arXiv},
      primaryClass={cs.CL}
}

@inproceedings{
he2021deberta,
title={DEBERTA: DECODING-ENHANCED BERT WITH DISENTANGLED ATTENTION},
author={Pengcheng He and Xiaodong Liu and Jianfeng Gao and Weizhu Chen},
booktitle={International Conference on Learning Representations},
year={2021},
url={https://openreview.net/forum?id=XPZIaotutsD}
}

@misc{schulman2017proximalpolicyoptimizationalgorithms,
      title={Proximal Policy Optimization Algorithms}, 
      author={John Schulman and Filip Wolski and Prafulla Dhariwal and Alec Radford and Oleg Klimov},
      year={2017},
      eprint={1707.06347},
      archivePrefix={arXiv},
      primaryClass={cs.LG},
      url={https://arxiv.org/abs/1707.06347}, 
}

@article{qwen3embedding,
  title={Qwen3 Embedding: Advancing Text Embedding and Reranking Through Foundation Models},
  author={Zhang, Yanzhao and Li, Mingxin and Long, Dingkun and Zhang, Xin and Lin, Huan and Yang, Baosong and Xie, Pengjun and Yang, An and Liu, Dayiheng and Lin, Junyang and Huang, Fei and Zhou, Jingren},
  journal={arXiv preprint arXiv:2506.05176},
  year={2025}
}

@article{zheng2023judging,
  title={Judging llm-as-a-judge with mt-bench and chatbot arena},
  author={Zheng, Lianmin and Chiang, Wei-Lin and Sheng, Ying and Zhuang, Siyuan and Wu, Zhanghao and Zhuang, Yonghao and Lin, Zi and Li, Zhuohan and Li, Dacheng and Xing, Eric and others},
  journal={Advances in neural information processing systems},
  volume={36},
  pages={46595--46623},
  year={2023}
}

@article{ouyang2022training,
  title={Training language models to follow instructions with human feedback},
  author={Ouyang, Long and Wu, Jeffrey and Jiang, Xu and Almeida, Diogo and Wainwright, Carroll and Mishkin, Pamela and Zhang, Chong and Agarwal, Sandhini and Slama, Katarina and Ray, Alex and others},
  journal={Advances in neural information processing systems},
  volume={35},
  pages={27730--27744},
  year={2022}
}

@article{williams1992simple,
  title={Simple statistical gradient-following algorithms for connectionist reinforcement learning},
  author={Williams, Ronald J},
  journal={Machine learning},
  volume={8},
  number={3},
  pages={229--256},
  year={1992},
  publisher={Springer}
}

@article{zheng2024sglang,
  title={Sglang: Efficient execution of structured language model programs},
  author={Zheng, Lianmin and Yin, Liangsheng and Xie, Zhiqiang and Sun, Chuyue and Huang, Jeff and Yu, Cody H and Cao, Shiyi and Kozyrakis, Christos and Stoica, Ion and Gonzalez, Joseph E and others},
  journal={Advances in neural information processing systems},
  volume={37},
  pages={62557--62583},
  year={2024}
}
\appendix
\appendix

\begin{figure*}[h]
  \centering
  \includegraphics[width=0.97\textwidth]{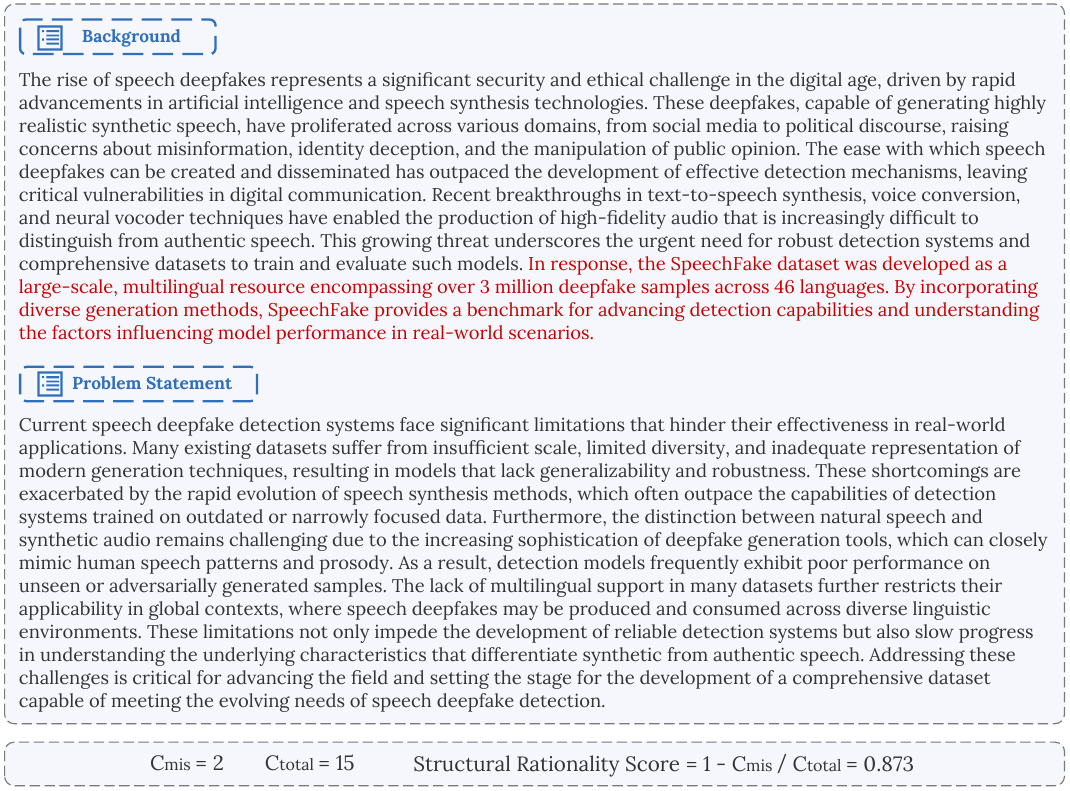}
  \caption{Illustration of the calculation of the structure score.}
  \label{fig:structure_case_appendix}
\end{figure*}
\section{Structural Rationality Illustration}
\label{app: structural rationality}

\paragraph{Auxiliary classifier.}
The Structural Rationality Score relies on an auxiliary classifier that predicts the intended rhetorical role of each sentence in a generated introduction. We use GPT-4o to annotate the rhetorical structure of approximately 1{,}000 introductions, drawn from both model-generated drafts and ACL Findings papers, and then train a DeBERTa-v3-large \citep{he2021debertav3,he2021deberta} classifier on the resulting sentence-level labels. The trained classifier achieves an accuracy of 94.76\% on a held-out test set, which is sufficient for providing reliable structural feedback during reinforcement learning.

\paragraph{Score computation.}
Given a generated paragraph $w_k$ assigned to a target rhetorical section (e.g., Background or Problem Statement), the classifier labels each constituent sentence with one of the four canonical sections. Sentences whose predicted label differs from the target are marked as structurally irrational. The Structural Rationality Score for $w_k$ is then computed as $r_{\text{struct}}(w_k)=1-C_{\text{mis}}/C_{\text{tot}}$, where $C_{\text{tot}}$ is the total number of sentences in $w_k$ and $C_{\text{mis}}$ is the number of mislabeled sentences. This formulation directly penalizes cross-stage content leakage and rewards rhetorical purity within each structural unit.

\paragraph{Illustrative example.}
Figure~\ref{fig:structure_case_appendix} visualizes the calculation process on two paragraphs generated by AutoSurvey, where the first paragraph is labeled as Background and the second as Problem Statement. Sentences highlighted in dark red are classified as structurally erroneous. Concretely, the Background paragraph contains 7 sentences, of which 2 prematurely describe methodological details, while the Problem Statement paragraph contains 8 sentences. Aggregating across both paragraphs yields a final Structural Rationality Score of $0.873$. The case also exposes a typical failure mode of multi-stage workflows: the model first states the problem, immediately proposes a dataset that resolves the issue, and then re-opens existing problems in the second paragraph, leading to repetitive mentions and structural confusion that StructPO is explicitly trained to avoid.

\section{Additional Experimental Details}
\label{app:additional_experimental_details}

This section provides additional implementation details that complement the experimental setup described in Section Experiments. These details include dataset splits, training configurations, baseline implementations, metric aggregation and token accounting.

\subsection{Dataset Splits and Construction}
\label{app:dataset_splits}

We construct the experimental corpus from approximately 3,200 ACL conference papers published between 2021 and 2025. PDF files are parsed into structured text with MinerU~\citep{wang2024mineruopensourcesolutionprecise}. For each paper, we extract the title, abstract, original introduction, figure captions, table captions and table contents. We also extract baseline-reference information from experimental sections when available.

The ACL 2025 papers are held out exclusively for testing, yielding 1,176 test instances. The remaining ACL papers are used for training and validation. Specifically, we use approximately 300 papers for supervised fine-tuning, around 1,550 papers for reinforcement learning training, and about 150 papers for validation during training. The held-out ACL 2025 set is not used in either SFT or RL training.

To obtain stage-level supervision, we decompose each original introduction into four rhetorical sections: Background, Problem and Limitations of Existing Methods, Brief Method Overview and Summary of Main Results, and Our Contributions. For each section, we further extract an outline. The decomposition is performed with GPT-4o~\citep{achiam2023gpt} using the prompt in Appendix~\ref{app:extract}. The resulting section-content and section-outline pairs are used for supervised training, reward computation and evaluation.

For out-of-domain evaluation, we additionally collect 141 unseen CVPR papers. These papers are used only for zero-shot transfer evaluation. No CVPR paper is used during SFT, RL training, reward-model construction or validation.

\begin{table}[h]
\centering
\footnotesize
\setlength{\tabcolsep}{3pt}

\label{tab:dataset_splits_appendix}
\begin{tabular*}{\columnwidth}{@{\extracolsep{\fill}}llcl@{}}
\toprule
\textbf{Split} & \textbf{Source} & \textbf{\#} & \textbf{Usage} \\
\midrule
SFT & ACL 21--24 & $\sim$300 & SFT \\
RL & ACL 21--24 & $\sim$1,550 & StructPO \\
Val. & ACL 21--24 & $\sim$150 & Validation \\
Test & ACL 2025 & 1,176 & Evaluation \\
OOD & CVPR & 141 & Transfer \\
\bottomrule
\end{tabular*}
\caption{Dataset splits used in our experiments.}
\end{table}

\subsection{Training Configuration}
\label{app:training_configuration}

StructPO is implemented with the Verl reinforcement learning framework~\citep{sheng2025hybridflow}, and SGLang~\citep{zheng2024sglang} is used for rollout and inference. We instantiate StructPO on Qwen2.5-7B-Instruct and Qwen3-8B~\citep{qwen2,yang2025qwen3technicalreport}. All reinforcement learning experiments use 8 H800 GPUs, a global batch size of 16 and a rollout size of 8.

For the dual-stream advantage fusion in Struct-aware Relative Advantage estimation, we set the global advantage weight to $\lambda=0.3$ and the local struct-aware advantage weight to $1-\lambda=0.7$. This setting places stronger emphasis on fine-grained stage-level control while still preserving whole-introduction semantic alignment. During RL training, we also apply a KL-divergence constraint to regularize the updated policy against excessive deviation from the reference model.

The length bandwidths follow the reward design: $\delta=6$ for section-level content length rewards and $\delta=12$ for the global length reward. During training, only tokens located within valid structural boundaries are optimized.

\begin{table}[h]
\centering

\label{tab:training_config_appendix}
\begin{tabular}{lc}
\toprule
\textbf{Configuration} & \textbf{Value} \\
\midrule
Training framework & Verl \\
Rollout / inference backend & SGLang \\
Hardware & 8 H800 GPUs \\
Global batch size & 16 \\
Rollout size & 8 \\
SFT data size & $\sim$300 papers \\
RL data size & $\sim$1,550 papers \\
Validation data size & $\sim$150 papers \\
Global advantage weight $\lambda$ & 0.3 \\
Local advantage weight $1-\lambda$ & 0.7 \\
Section length bandwidth & 6 \\
Global length bandwidth & 12 \\
KL regularization & Enabled \\
\bottomrule
\end{tabular}
\caption{Main training configurations for StructPO.}
\end{table}

\subsection{Baseline Implementation Details}
\label{app:baseline_implementation_details}

We compare StructPO with prompt-based, workflow-based, supervised stage-token and closed-source LLM baselines.

\paragraph{Pure Prompt.}
The model directly generates the introduction from the title, abstract, figure information, table information and baseline-reference information. No explicit outline guidance, stage tokens or refinement loop is used.

\paragraph{ELABORATE Prompting.}
We adapt ELABORATE prompting~\citep{garg2025let} to introduction generation by enforcing a four-part rhetorical structure covering context, research gap, proposed solution and contribution. Unlike StructPO, this baseline relies only on prompt-level structure control and does not update the model parameters.

\paragraph{AutoSurvey.}
AutoSurvey~\citep{NEURIPS2024_d07a9fc7} is adapted as a workflow-based baseline. The original survey-generation workflow is modified for introduction generation. The system first produces structured outlines and then expands them into dense introduction content. This baseline represents explicit multi-stage orchestration.

\paragraph{SurveyForge.}
Same as AutoSurvey, SurveyForge is also adapted as a workflow-based baseline.

\paragraph{STIG.}
STIG~\citep{zhang2025eliminating} uses the same stage-token format as StructPO, but it is trained only with supervised fine-tuning. It does not use Struct-aware Relative Advantage estimation, global-local reward fusion or refinement-guided optimization. This baseline isolates the effect of stage-token supervision without reinforcement learning.

\paragraph{Refinement without Training.}
This baseline performs inference-time refinement using the SFT model. The model first generates an initial draft and then revises it with the refinement prompt in Appendix~\ref{app:prompt_structpo}. Unlike StructPO, the revision behavior is not distilled into the policy during training, so the method requires an additional inference pass.

\paragraph{Closed-source LLMs.}
We evaluate GPT-4o and GPT-5.1 through API access. They are prompted with the same structured input materials used by StructPO, including the title, abstract, figure information, table information and baseline-reference information. These models serve as strong closed-source references for semantic quality, writing fluency and factual consistency.

\subsection{Token Accounting Protocol}
\label{app:token_accounting}

For the token-efficiency analysis in Section Analysis, we separate inference-time token consumption into \textit{overhead tokens} and \textit{effective tokens}. Overhead tokens include input context, system prompts, formatting instructions, intermediate outlines, intermediate drafts, revision instructions and other workflow states that are consumed during generation but are not part of the final delivered introduction. Effective tokens refer to the final introduction content presented to the user after removing structural markers and intermediate workflow artifacts.

For workflow-based systems such as AutoSurvey, token usage is accumulated across all intermediate calls. Therefore, repeated context replay and intermediate outputs are counted each time they are consumed by the workflow. For single-pass systems such as StructPO and STIG, the token count is computed from the single inference call. The effective rate is defined as:
\begin{equation}
\small
\text{Effective Rate}
=
\frac{\text{Effective Tokens}}
{\text{Effective Tokens}+\text{Overhead Tokens}}.
\end{equation}
This protocol allows us to compare multi-stage workflows and single-pass policies under the same accounting framework.

\begin{figure*}[t!]
  \centering
  \includegraphics[width=0.97\textwidth]{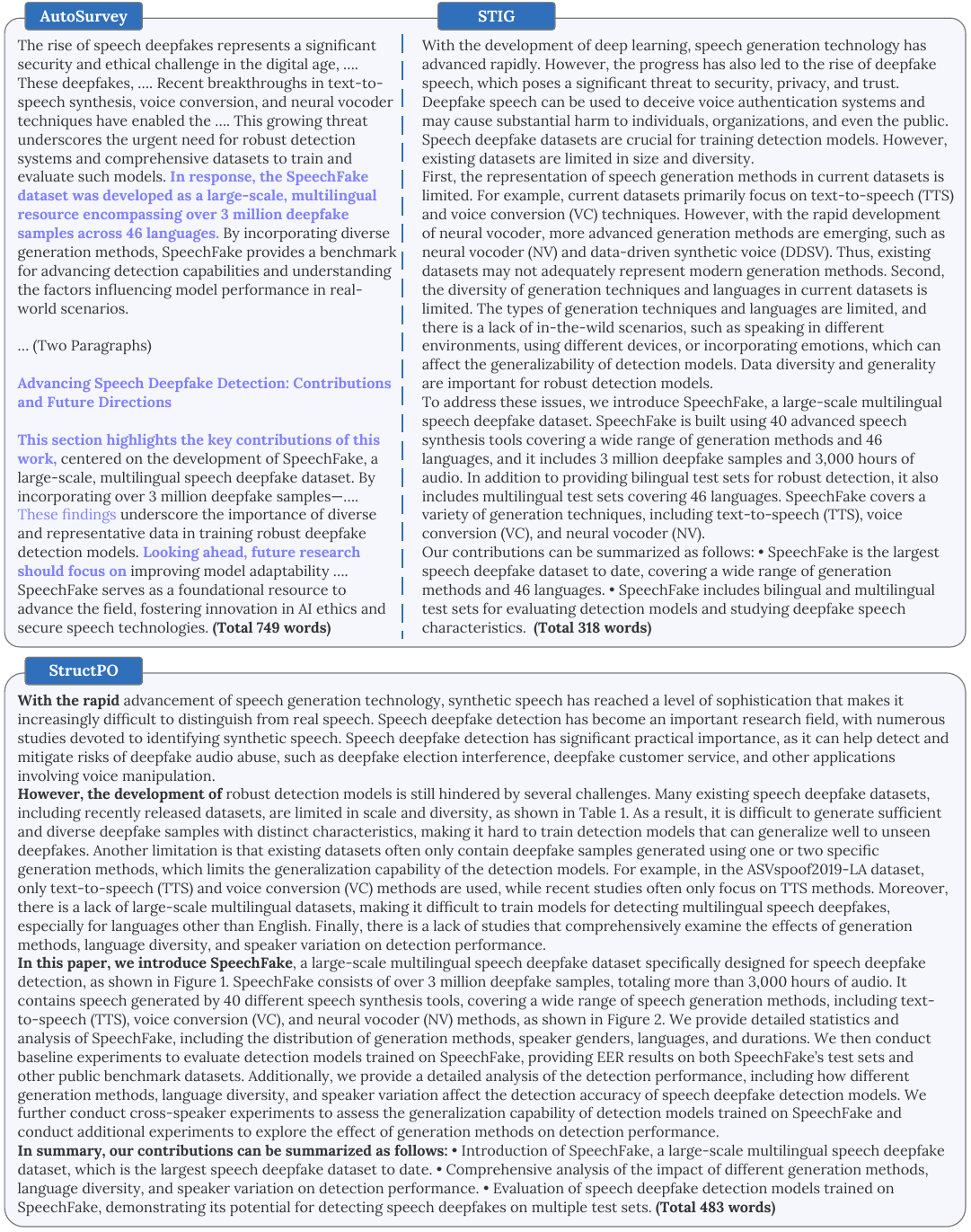}
  \caption{Qualitative comparison among AutoSurvey, STIG and StructPO on a real ACL test case \citep{huang-etal-2025-speechfake}. Blue marks structurally inappropriate wording in AutoSurvey, the bottom annotation reports the insufficient total word count of STIG, and bold transitional anchors highlight the rhetorical bridges produced by StructPO.}
  \label{fig:case_study}
\end{figure*}
\section{Qualitative Test Case}
\label{app:case}

To complement the brief discussion in the main text, we provide a more detailed qualitative diagnosis of the three competing paradigms on a representative ACL 2025 test paper, \textit{SpeechFake} \citep{huang-etal-2025-speechfake}. Figure~\ref{fig:case_study} visualizes the generated introductions side by side.

\paragraph{AutoSurvey: cross-stage drift and meta-commentary.}
Because AutoSurvey decomposes generation into independent agent calls, the resulting text exhibits clear cross-stage drift. As highlighted in blue, the model prematurely reveals the proposed dataset (``the SpeechFake dataset was developed...'') at the end of the very first background paragraph, breaking the rhetorical suspense expected in an academic introduction. In the contribution section, the agent further hallucinates a markdown-style subtitle and inserts unnatural meta-commentary, treating each stage as a self-contained blog post rather than a coherent component of a unified introduction. These artifacts illustrate how external multi-stage orchestration weakens the macroscopic narrative when individual agents lose access to the document-level context.

\paragraph{STIG: length collapse and shallow synthesis.}
STIG successfully partitions the output into a four-paragraph skeleton through stage-token supervision, but suffers from the well-known length-collapse failure mode of pure SFT. The full introduction contains only 318 words, which is markedly insufficient for a top-tier conference paper. The second paragraph repeatedly uses the word ``limited'', indicating shallow synthesis, and the final contribution bullets are essentially copy-pasted from the preceding method paragraph rather than abstracted at a higher level. This confirms that behavioral cloning alone, even with explicit stage tokens, is unable to encourage rich and well-elaborated academic content.

\paragraph{StructPO: balanced structure and dense content.}
In contrast, StructPO produces a balanced introduction that follows the background--problem--method--contribution progression with explicit rhetorical anchors (e.g., ``With the rapid advancement...'', ``However...'', ``In this paper...'', ``In summary...''), as marked in bold. Unlike AutoSurvey, StructPO avoids cross-stage drift; unlike STIG, it generates dense, paper-specific content, referencing concrete artifacts such as Table~1 and Figure~2, and reporting specific evaluation metrics including equal error rate (EER) and cross-speaker results. The contribution section is condensed into three well-formatted bullet points that synthesize, rather than duplicate, the preceding paragraphs. This case suggests that struct-aware policy optimization can internalize the workflow-style organization of writing while preserving coherent single-pass generation.

\section{Human Evaluation Details}
\label{app:human_eval}

We provide additional details of the blind human evaluation. We randomly sample $N=30$ papers from the ACL test set. For each paper, we present the source abstract and two generated introductions to $E=3$ human evaluators with NLP research experience. The two introductions are generated by StructPO and GPT-5.1, respectively. Model identities are hidden from the evaluators and the presentation order is randomized to reduce position bias. Evaluators are instructed to choose the better introduction based on a holistic assessment of logical coherence, structural completeness and academic writing quality.

For each paper $i \in \{1,\dots,N\}$ and evaluator $j \in \{1,\dots,E\}$, we define a binary preference variable $v_{i,j}^{(M)} \in \{0,1\}$, where $v_{i,j}^{(M)}=1$ if evaluator $j$ prefers model $M \in \{\text{StructPO}, \text{GPT-5.1}\}$ for paper $i$ and $0$ otherwise. We compute two metrics. \textbf{Win Votes} measures the overall vote share across all annotations:
\begin{equation}
    \text{WinVotes}^{(M)} =
    \frac{1}{N \cdot E}
    \sum_{i=1}^{N}
    \sum_{j=1}^{E}
    v_{i,j}^{(M)}
    \times 100\%.
\end{equation}

\textbf{Win Rate} measures the proportion of papers for which a model wins by majority vote:
\begin{equation}
    \text{WinRate}^{(M)} =
    \frac{1}{N}
    \sum_{i=1}^{N}
    \mathbb{I}
    \left(
    \sum_{j=1}^{E}
    v_{i,j}^{(M)}
    >
    \frac{E}{2}
    \right)
    \times 100\%.
\end{equation}

Table~\ref{tab:human_annotation} reports the raw annotation results for all 30 sampled papers. ``StructPO'' and ``GPT'' indicate the model preferred by each annotator and the final winner is determined by majority vote.

\begin{table}[t]
\centering
\begin{tabular}{lllll}
\toprule
\textbf{Paper ID} & \textbf{Human 1} & \textbf{Human 2} & \textbf{Human 3} & \textbf{Winner} \\
\midrule
1 & StructPO & GPT & StructPO & StructPO \\
104 & GPT & StructPO & GPT & GPT \\
106 & StructPO & GPT & GPT & GPT \\
109 & StructPO & StructPO & StructPO & StructPO \\
110 & GPT & GPT & StructPO & GPT \\
1002 & GPT & GPT & StructPO & GPT \\
1005 & GPT & GPT & StructPO & GPT \\
1007 & GPT & StructPO & GPT & GPT \\
1009 & GPT & StructPO & StructPO & StructPO \\
1020 & StructPO & StructPO & GPT & StructPO \\
1025 & GPT & GPT & GPT & GPT \\
1029 & StructPO & GPT & StructPO & StructPO \\
1032 & GPT & GPT & StructPO & GPT \\
1033 & StructPO & GPT & StructPO & StructPO \\
1035 & GPT & StructPO & StructPO & StructPO \\
1036 & StructPO & StructPO & GPT & StructPO \\
1037 & StructPO & GPT & StructPO & StructPO \\
1044 & StructPO & GPT & GPT & GPT \\
1045 & StructPO & GPT & StructPO & StructPO \\
1047 & StructPO & StructPO & StructPO & StructPO \\
1058 & StructPO & GPT & GPT & GPT \\
1068 & GPT & StructPO & StructPO & StructPO \\
1069 & StructPO & StructPO & GPT & StructPO \\
1072 & StructPO & StructPO & StructPO & StructPO \\
1081 & GPT & StructPO & GPT & GPT \\
1089 & StructPO & StructPO & StructPO & StructPO \\
1107 & StructPO & GPT & GPT & GPT \\
1122 & GPT & GPT & StructPO & GPT \\
1123 & StructPO & StructPO & StructPO & StructPO \\
1124 & StructPO & GPT & GPT & GPT \\
\bottomrule
\end{tabular}
\caption{Raw human-annotation data for 30 sampled papers.}
\label{tab:human_annotation}
\end{table}

\section{Prompt for Structural Decomposition}
\label{app:extract}

The following prompt is used to decompose ACL introductions into section-level content and outlines.

\begin{lstlisting}
Please break down the introduction section of the following academic paper into a structured outline format for academic discussion purposes.
The content should be divided into the following four sections, extracting key points for each:

1. Background: Basic background and significance of the research field
   - Number of points: 2-4
2. Problem and Limitations of Existing Methods: Current issues, challenges and limitations of existing methods
   - Number of points: 2-6
3. Brief Method Overview and Summary of Main Results: Overview of the proposed method, main experimental results and findings
   - Number of points: 4-8
4. Our Contributions: Main contributions and innovations of the paper
   - Number of points: 2-3

Please output in the following JSON format, including outline points and paragraphs classified by section:

{
    "sections": {
        "Background": "Combine all paragraphs and sentences belonging to the background section",
        "Problem and Limitations of Existing Methods": "Combine all paragraphs and sentences belonging to the problems and limitations section",
        "Brief Method Overview and Summary of Main Results": "Combine all paragraphs and sentences belonging to the method overview and main results sections",
        "Our Contributions": "Combine all paragraphs and sentences belonging to the contributions section"
    },
    "outline": {
        "Background": ["Point 1", "Point 2", "..."],
        "Problem and Limitations of Existing Methods": ["Point 1", "Point 2", "..."],
        "Brief Method Overview and Summary of Main Results": ["Point 1", "Point 2", "..."],
        "Our Contributions": ["Point 1", "Point 2", "..."]
    }
}

Introduction content:
{text}

Notes:
1. Analyze the content coherently and categorize it into the appropriate sections, strictly controlling the number of points for each section.
2. When assigning sections, ensure continuity; for example, Background must be at the beginning of the article and if there is an Our Contributions section, it must be at the end. There should be no section 1, section 2, section 1 sequences.
3. Some papers may not have a section similar to Our Contributions; if so, generate an empty Our Contributions field.
4. First, divide the sections, then perform an outline analysis to identify key points.
5. Do not use demonstrative pronouns like "this" or "the model" in the key points; use specific names if available.
\end{lstlisting}

\section{Inference and Refinement Prompts}
\label{app:prompt_structpo}

The following prompts are used for inference and refinement.

\begin{lstlisting}
As an academic writing expert who has completed a research project and is currently in the paper-writing stage, please draft the introduction section based on the available materials.
Follow the format of Outline to Content, first drafting the outline of this section and then the content.
When writing, ensure logical coherence and smooth transitions and use fluent and standard academic English.

Style and Content Requirements:
- Maintain a formal academic tone.
- Be as coherent and concise as possible and directly related to the title and abstract.
- Use transitional phrases effectively.

Citation Instructions:
- Do not mention any citations. For example, "(Smith et al.)".
- Do not use reference formats such as \ref.

Compose the Introduction of an ACL paper based on the corresponding research materials. For each sub-section, first list the outline and then write the corresponding content of that section. You need to write four sections:
1. Background: Provide the research background and the current status of the field. (The content text is about 90 words)
2. Problem and Limitations of Existing Methods: Describe the research problem and the limitations of existing methods. (The content text is about 180 words)
3. Brief Method Overview and Summary of Main Results: Briefly introduce the proposed method and summarize the main results. (The content text is about 230 words)
4. Our Contributions: Summarize the contributions of this paper. (The content text is about 70 words)

Please write introduction with the following writing format and use <STAGE> and <END> markers to represent writing stages:
<STAGE0> Outline for Background:
(Outline for Background) <END0>
<STAGE1> Contents for Background:
(Contents for Background) <END1>
<STAGE2> Outline for Problem and Limitations of Existing Methods:
(Outline for Problem and Limitations of Existing Methods) <END2>
<STAGE3> Contents for Problem and Limitations of Existing Methods:
(Contents for Problem and Limitations of Existing Methods) <END3>
<STAGE4> Outline for Brief Method Overview and Summary of Main Results:
(Outline for Brief Method Overview and Summary of Main Results) <END4>
<STAGE5> Contents for Brief Method Overview and Summary of Main Results:
(Contents for Brief Method Overview and Summary of Main Results) <END5>
<STAGE6> Outline for Our Contributions:
(Outline for Our Contributions) <END6>
<STAGE7> Contents for Our Contributions:
(Contents for Our Contributions) <END7>

Research materials:
Title: {data['title']}
Abstract: {data['abstract']}
Figure information: {data['figures']}
Table information: {data['tables']}
Baseline references: {data['ref']}
\end{lstlisting}

\begin{lstlisting}
Please carefully review your writing according to revision suggestions and make revisions to improve logical coherence, clarity of expression and adherence to academic standards. Follow the original writing format and provide the revised content directly.
\end{lstlisting}

\section{Evaluation Prompts for AWQ and SFC}
\label{app:metrics}

The following prompts are used for AWQ and SFC evaluation.

\begin{lstlisting}
system_prompt = """You are a senior academic writing expert evaluating the writing quality of a research paper Introduction, particularly for CS/NLP venues (e.g., ACL, EMNLP, NAACL).

CRITICAL INSTRUCTION: 
- Evaluate ONLY the writing quality, structure, and academic tone.
- COMPLETELY IGNORE all citations and references.
- Do NOT consider whether citations exist, are correct, or are properly formatted.
- Treat the text as if all citation markers (e.g., "[1]", "(Author, Year)") do not exist.
- Focus purely on: vocabulary, sentence structure, logical flow, content balance, and professional expression."""

user_prompt = f"""Generated Introduction:
{generated_introduction}

### Evaluation Criteria (IGNORE ALL CITATIONS)

1. Academic Vocabulary & Formal Tone
   - Uses precise, formal terminology appropriate for scholarly writing
   - Avoids colloquial expressions, slang, or overly casual language
   - Employs appropriate hedging (e.g., "may suggest", "appears to", "potentially")
   - Uses domain-specific technical terms correctly
   - Avoids vague expressions (e.g., "many studies", "significant improvement" without specifics)
   - GOOD: Using concrete numbers and specific comparisons (e.g., "10 times larger", "15% improvement")
   - BAD: Empty modifiers without support (e.g., "pioneering", "comprehensive", "thorough" without evidence)

2. Logical Structure & Argumentation
   - Clear problem statement and research motivation
   - Logical progression: Background - Research Gap - Proposed Solution - Contributions
   - Well-organized paragraphs with clear topic sentences
   - Smooth transitions between ideas and sections
   - Coherent narrative that guides the reader
   
   CRITICAL - Paragraph Balance:
   - Paragraphs should have UNEQUAL lengths reflecting their importance
   - Core contributions/methods should be MORE prominent (longer/more detailed)
   - Background should be CONCISE (not a lengthy tutorial)
   - RED FLAG: All paragraphs having similar length (~equal words) indicates poor prioritization
   
   What to PENALIZE (max 2 points):
   - Mechanical section preview: "Section 3 presents..., Section 4 describes..., Section 5 shows..."
   - Method/experiment paragraphs as task lists: "We do A. We then do B. We further do C. Additionally, we do D."
   - Repetitive content across paragraphs (e.g., same limitation stated twice)
   
   What is ACCEPTABLE (especially in CS/NLP):
   - Contributions listed with bullet points or numbered items (e.g., " First, ...  Second, ...")
   - "Our contributions are as follows: (1)... (2)... (3)..."
   - This is STANDARD practice in ACL, EMNLP, NAACL papers - do NOT penalize

3. Content Proportion & Emphasis
   - Background: Should be CONCISE (ideally 15-25% of introduction), NOT a textbook tutorial
   - Research Gap: Clearly and specifically articulated with concrete examples or evidence
   - Proposed Solution: Should be prominent, with specific details (numbers, comparisons)
   - Contributions: Should be SPECIFIC and CONCRETE, not vague claims
   
   Balance Guidelines:
   - Background > 40% - likely too long
   - Contributions < 15% - likely too brief or weak
   - All paragraphs about 10% of each other - poor structure (lack of emphasis)
   
   Good Contributions Example:
   "We introduce X, comprising 3M samples across 46 languages-10 times larger than prior benchmarks."
   
   Bad Contributions Example:
   "We introduce X, a pioneering and comprehensive dataset for the research community."

4. Sentence Quality & Readability
   - Varied sentence structures (not repetitive patterns)
   - Appropriate sentence length (not too long or choppy)
   - Clear and unambiguous expressions
   - Proper grammar and syntax
   - No awkward phrasing or unclear constructions
   - PENALIZE: Repetitive sentence patterns (e.g., "as shown in Figure X" used 3+ times)
   - PENALIZE: Consecutive "We + verb" sentences in non-contribution sections

5. Introduction-Specific Requirements
   - Opening: Engaging first paragraph that establishes importance and relevance
   - BAD Opening: "With the rapid advancement of X..." / "In recent years, X has attracted..." 
   - GOOD Opening: Start with a concrete problem, striking fact, or specific context
   - Closing: Clear statement of contributions (bullet points acceptable in CS/NLP)
   - Self-contained: Reader understands the paper's purpose without needing other sections
   - References to Tables/Figures: Good introductions often reference Table 1 or Figure 1 for support

6. Professional Academic Style
   - Appropriate use of active/passive voice
   - Objective and impersonal tone where appropriate
   - Concise writing without unnecessary redundancy
   - PENALIZE: Redundant phrases like "contributions can be summarized as follows" (just say "Our contributions:")

### Common Defects Checklist
- [ ] Excessive background (>40% of introduction)
- [ ] All paragraphs roughly equal length (no emphasis)
- [ ] Vague or missing problem statement
- [ ] Generic contributions without specific numbers/findings
- [ ] "Firstly/Secondly/Thirdly/Lastly" mechanical listing (worse than bullet points)
- [ ] Repetitive content between paragraphs
- [ ] Method paragraph as task list ("We do A. We then do B...")
- [ ] Normal opening ("With the rapid advancement...")
- [ ] No reference to Tables/Figures for evidence
- [ ] Empty modifiers ("comprehensive", "thorough", "pioneering") without support

### Scoring Guide (1-5)
- 5: Excellent. Publication-ready for top venues. Well-balanced content with clear emphasis on contributions, engaging opening, logical flow, specific claims with evidence, appropriate length. Contributions are concrete with numbers/findings.
- 4: Good. Solid academic writing suitable for submission. Minor issues: slightly unbalanced content, some repetitive patterns, or contributions could be more specific. References Tables/Figures appropriately.
- 3: Acceptable. Recognizably academic but has noticeable problems: somewhat imbalanced paragraphs, normal opening, vague contributions, or some awkward sentences. Needs revision before submission.
- 2: Poor. Significant issues: clearly imbalanced (e.g., 50% background), mechanical section listing, task-list method paragraphs, repetitive content, empty claims. Needs substantial revision.
- 1: Unacceptable. Severely flawed: no clear structure, mechanical listing throughout, disorganized, or unprofessional writing.

### Output Format
{{"vocabulary": "<brief comment>", "structure": "<brief comment, note paragraph balance>", "readability": "<brief comment>", "overall": "<overall assessment with main issues>", "score": <integer 1-5>}}"""
\end{lstlisting}

\begin{lstlisting}
system_prompt = """You are a lenient academic reviewer checking for hallucinations.

ONE STRICT RULE: Fabricated citations are unacceptable (Score 1-2).

EVERYTHING ELSE: Be lenient. If content can be reasonably inferred from source materials or is common academic knowledge, it is NOT hallucination.

Academic introductions naturally expand beyond abstracts. This is expected, not hallucination.

Respond with JSON only."""

user_prompt = f"""Source Abstract:
{abstract}

Source Table Information:
{table_info}

Source Figure Captions:
{figure_info}

Generated Introduction:
{generated_introduction}

### Evaluation Criteria

STRICT: Fake Citations (Zero Tolerance)
- Any citation "(Author, Year)", "[1]", etc. MUST exist in sources
- Fake citation found - Score 1-2

LENIENT: Everything Else
Accept as valid if content:
- Can be inferred or derived from source materials
- Is general background or domain knowledge
- Is reasonable elaboration or explanation
- Is logical implication of what sources describe
- Uses approximate numbers (e.g., "about 3M" = "3,000,000")

Only flag as hallucination if:
- Directly contradicts source materials
- Completely fabricates results/methods with no basis in sources

### Scoring Guide (1-5)

- 5: Good - No fake citations. Content aligns with or is inferable from sources.

- 4: Fine - No fake citations. Reasonable expansions. Perhaps minor liberties but acceptable.

- 3: Borderline - No fake citations. Some content seems loosely connected to sources but not contradictory.

- 2: Problematic - Contains fake citation(s), OR directly contradicts sources.

- 1: Severe - Multiple fake citations, OR fabricates core content entirely.

Default Assumption: If no fake citations and no obvious contradictions - Score 4-5

### Output Format
{{"fake_citations": "<citations NOT in sources, or 'None'>", "factuality_issues": "<only obvious contradictions/fabrications, or 'None'>", "score": <integer 1-5>}}"""
\end{lstlisting}

\end{document}